\documentclass[conference]{IEEEtran}
\IEEEoverridecommandlockouts

\newif\ifreviewnumbers
\reviewnumbersfalse

\usepackage[switch]{lineno}
\usepackage{etoolbox}
\usepackage{xcolor}
\AtBeginEnvironment{figure}{\ifreviewnumbers\internallinenumbers\fi}
\AtBeginEnvironment{table}{\ifreviewnumbers\internallinenumbers\fi}

\usepackage[colorlinks=true,linkcolor=black,citecolor=black,urlcolor=blue]{hyperref}
\usepackage{cite}
\usepackage{balance}
\usepackage{amsmath,amssymb,amsfonts}
\usepackage{algorithmic}
\usepackage{graphicx}
\usepackage{textcomp}
\usepackage{booktabs}
\usepackage[table]{xcolor}
\usepackage{colortbl}
\usepackage{tikz}
\usepackage{float}
\usepackage{subcaption}
\usetikzlibrary{spy}

\definecolor{posStrong}{HTML}{1B7F3A}
\definecolor{posMed}{HTML}{4CAF50}
\definecolor{posWeak}{HTML}{A5D6A7}
\definecolor{negWeak}{HTML}{EF9A9A}
\definecolor{negMed}{HTML}{E53935}
\definecolor{negStrong}{HTML}{B71C1C}

\newcommand{\best}[1]{\textbf{#1}}
\newcommand{\runnerup}[1]{\underline{#1}}
\newcommand{\cellps}[2]{#1\,\textcolor{posStrong}{\scriptsize$\uparrow$#2\%}}
\newcommand{\cellpm}[2]{#1\,\textcolor{posMed}{\scriptsize$\uparrow$#2\%}}
\newcommand{\cellpw}[2]{#1\,\textcolor{posWeak}{\scriptsize$\uparrow$#2\%}}
\newcommand{\cellnw}[2]{#1\,\textcolor{negWeak}{\scriptsize$\downarrow$#2\%}}
\newcommand{\cellnm}[2]{#1\,\textcolor{negMed}{\scriptsize$\downarrow$#2\%}}
\newcommand{\cellns}[2]{#1\,\textcolor{negStrong}{\scriptsize$\downarrow$#2\%}}

\begin{document}
\pagestyle{empty}

\title{From Benchmark Performance\\to Tool Deployment:\\Human-in-the-Loop Anomaly Detection}
\author{
\IEEEauthorblockN{
Mike Szklarzewski\IEEEauthorrefmark{1},
CJ George\IEEEauthorrefmark{1},
Gavin Smithson\IEEEauthorrefmark{1},
Christopher Stokes\IEEEauthorrefmark{1},
Dakota Fulp\IEEEauthorrefmark{1},
William M. Jones\IEEEauthorrefmark{1}\IEEEauthorrefmark{4}\\
Benjamin Wynn\IEEEauthorrefmark{2},
Alexander Ur\IEEEauthorrefmark{3},
Agit Yesiloz\IEEEauthorrefmark{5},
Clint Kallenbach\IEEEauthorrefmark{6}\\
Mark Swartz\IEEEauthorrefmark{6},
Nathan DeBardeleben\IEEEauthorrefmark{4},
Sharmistha Chakrabarti\IEEEauthorrefmark{4}
}

\IEEEauthorblockA{
\IEEEauthorrefmark{1} Department of Computing Sciences, Coastal Carolina University, Conway, SC, USA\\
\IEEEauthorrefmark{2} Academy of Arts Science and Technology, Horry County School District, Myrtle Beach, SC, USA\\
\IEEEauthorrefmark{3} Hillcrest High School, Greenville County School District, Simpsonville, SC, USA\\
\IEEEauthorrefmark{4} High Performance Computing Design, Los Alamos National Laboratory, Los Alamos, NM, USA\\
\IEEEauthorrefmark{5} Operational Technology Center of Excellence, Savannah River Nuclear Solutions, LLC, Aiken, SC, USA\\
\IEEEauthorrefmark{6} Production Process Engineering, Los Alamos National Laboratory, Los Alamos, NM, USA
}
}

\maketitle
\thispagestyle{empty}

\begingroup
\renewcommand\thefootnote{}
\makeatletter
\long\def\@makefntext#1{\parindent 0pt\noindent#1}
\makeatother
\footnotetext{\footnotesize%
\emph{Corresponding author: Sharmistha Chakrabarti}
(\href{mailto:schakrabarti@lanl.gov}{schakrabarti@lanl.gov}).
\space This manuscript has been approved for unlimited release and has
been assigned LA-UR-26-23692. This work has been authored by an
employee of Triad National Security, LLC which operates Los Alamos
National Laboratory under Contract No. 89233218CNA000001 with the
U.S. Department of Energy/National Nuclear Security Administration.
The publisher, by accepting the article for publication, acknowledges
that the United States Government retains a non-exclusive, paid-up,
irrevocable, world-wide license to publish or reproduce the published
form of the manuscript, or allow others to do so, for United States
Government purposes.%
}
\endgroup

\begin{abstract}

Automated anomaly detection methods often report strong performance
on curated academic benchmarks, but their behavior under real-world
industrial conditions is less clear. In this work, we evaluate 19
unsupervised anomaly detection models on the BowTie dataset, a
challenging manufacturing dataset with reflective surfaces, subtle
defects, and profile-specific variation. In contrast to benchmark
results, we observe that model performance is less stable than
typically reported on standard benchmarks such as MVTec AD, highly
sensitive to preprocessing, and inconsistent across conditions,
with no single approach emerging as uniformly robust; a consensus
audit further indicates that nominal-data quality affects deployment.

Motivated by these findings, we developed and initially deployed a unified
human-in-the-loop framework for manufactured-part inspection that
combines image annotation, AI-assisted defect detection, and an
integrated validation engine, replacing a prior manual visual
inspection and documentation workflow. The system supports
heatmap-guided defect review,
SAM-refined candidate regions for inspector acceptance, rejection, or
boundary adjustment, mask evaluation where annotations exist, and
review history for inspector consistency and onboarding.
Together, the results highlight the gap
between benchmark performance and deployment reality, and provide
a practical framework for addressing it.

\end{abstract}


\section{Introduction}
Modern manufacturing and quality control systems continue to grow in scale and
complexity, placing increasing demands on performance, efficiency, and
reliability. Within
national laboratory environments such as Los Alamos
National Laboratory (LANL),
these challenges are further amplified by the need
to design, fabricate, and evaluate highly specialized, often one-off or
low-volume components that directly support
the national security mission.
Unlike traditional high-volume manufacturing, these workflows frequently
involve custom geometries, evolving design requirements,
and strict engineering
tolerances, where even subtle defects can have significant downstream
implications.

These challenges are particularly evident in scenarios where defects are rare,
subtle, or highly variable in morphology, such as in precision manufacturing
and experimental system evaluation.
In these environments
(the
\textit{BowTie} dataset \cite{oceans11} is one such example),
identifying the mere presence of an anomaly is
insufficient; researchers and inspectors must precisely localize and quantify
the spatial extent of the defect to ensure structural integrity and compliance
with strict engineering tolerances.

However, deploying anomaly detection models in a practical, operational
workflow reveals systemic bottlenecks. Many existing solutions are tightly
coupled to command-line interfaces or ``black box'' inference engines. This
creates a persistent operational disconnect between the \textit{data annotation}
process (curating ground-truth), the \textit{model inference} stage (generating
and visualizing predictions), and \textit{statistical validation}
(quantifying model success). In the motivating manual visual inspection and
documentation workflow, inspectors
manually reviewed each part image in a standalone viewer and logged defect
observations and accept/reject decisions in a spreadsheet. That
process relied almost entirely on individual visual judgment, without
preserving localized defect rationale or supporting consistent comparison
across inspectors. Our goal was therefore twofold: to make inspection easier
and more consistent by combining image viewing, defect annotation, and
ML-assisted anomaly localization in one environment while keeping final
disposition under inspector control, and to create a continuous improvement
cycle in which finalized inspections become structured data for retraining and
redeploying models over time. This tighter integration also improves
downstream communication, reproducibility, and quantitative evaluation, while
preserving review history for inspector onboarding and consistency
checks against expert-reviewed cases.

This paper contributes a 19-model BowTie benchmark under profile and preprocessing
variation, a consensus audit of nominal-data quality, and a unified, extensible
software framework comprising \textit{AnnoMate} for guided image annotation and
review, \textit{MicroSentryAI} for AI-assisted defect localization and
segmentation review, and a dedicated \textit{Validation Engine} for mask-based
evaluation and comparison against expert reference annotations where ground
truth is available \cite{annomate-microsentryai-workflow}.
The same reference-based comparison also supports inspector onboarding by
allowing trainee decisions and annotations to be compared against
expert-reviewed cases. The framework keeps the modules decoupled but
interoperable; Section~\ref{sec:tool_overview} gives the architecture and
workflow details. To keep that motivation concrete, the paper first establishes
the BowTie study design (Section~\ref{sec:experimentation}), empirical findings
(Sections~\ref{sec:results}--\ref{sec:label_noise}), and then returns to the
unified framework that operationalizes those lessons in practice
(Section~\ref{sec:tool_overview}).

\section{Related Work}
\label{sec:related}

Most prior work in industrial anomaly detection (IAD) has focused on model accuracy, sensing pipelines, and automated monitoring rather than on inspector-centered annotation workflows. Reviews of anomaly detection in additive manufacturing and practical inspection systems for metallic parts or laser directed energy deposition emphasize process monitoring, defect localization, and deployment-oriented sensing \cite{SAHAR2023100803,BENMOUSSAT201368,KAJI2022624}. At the annotation level, spatial ground truth is often created with general-purpose tools such as VIA and LabelMe \cite{dutta2019via,russell2008labelme}. These tools are effective polygon editors, but they are not tightly coupled to anomaly-model inference, interactive threshold tuning, contour simplification, synchronized heatmap visualization, or reciprocal comparison between human annotations and model outputs. AnnoMate is intended to address that gap.

The broader anomaly-detection literature provides the conceptual basis for MicroSentryAI. Foundational surveys organize anomaly detection around recurring methodological families and highlight challenges such as limited labels, high-dimensional data, low recall, and anomaly explanation \cite{chandola2009anomaly,pang2021anomaly}. Anomalib provides a common algorithmic substrate for unsupervised anomaly detection (UAD) by packaging state-of-the-art models together with training, inference, benchmarking, visualization, and hyperparameter tuning \cite{akcay2022anomalib}. Our contribution differs in that it keeps model execution inside an inspector-facing environment for interactive heatmap visualization, threshold tuning, contour simplification, and direct transfer of AI-generated masks back into the annotation pipeline. This emphasis is further motivated by evidence that preprocessing benefits are strongly method-dependent rather than universal \cite{zhang2023makesgooddataaugmentation}.

Evaluation in anomaly detection has progressively shifted from image-level classification toward fine-grained spatial localization. Benchmarks such as MVTec AD and VisA enable this shift by providing pixel-level annotations and overlap-based metrics, but they also assume that the original mask is a stable representation of the defect \cite{bergmann2021mvtec}. In real industrial settings, expert annotators may agree on the presence and location of a defect while differing on its precise boundary. For that reason, IoU, precision, and recall remain necessary but not always sufficient, and localization-oriented measurements such as centroid distance provide a complementary view of agreement. Our evaluation engine follows that line of work: it serves not only as a scoreboard for ranking models, but also as a calibration layer for comparison against a shared, reproducible spatial definition of a defect.

\section{Experimentation}
\label{sec:experimentation}

\subsection{Dataset and Topologies}
\label{subsec:dataset}
\begin{figure}[!t]
    \centering
    \includegraphics[width=\linewidth]{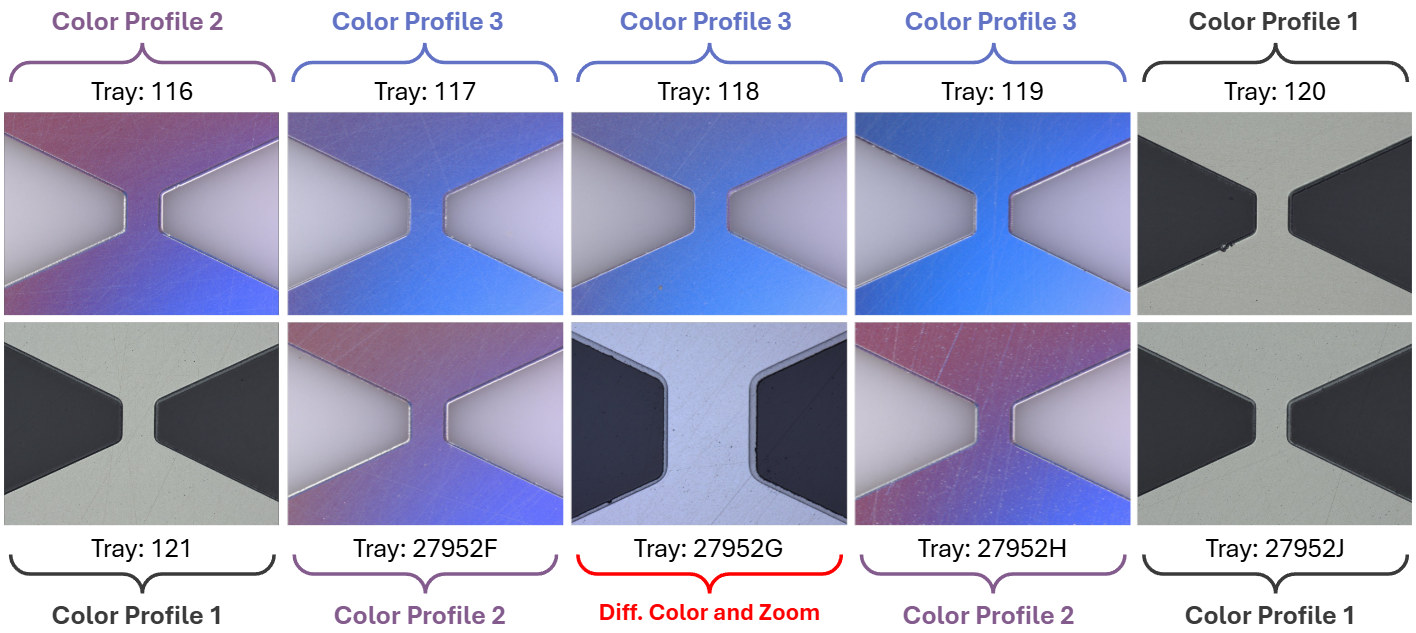}
    \caption{Representative tray images for CP1–CP3, with excluded tray 27952G shown for contrast. Tray 27952G is excluded from the benchmark because its color profile and zoom factor are inconsistent with the included trays, and it was later discarded by inspectors.}
    \label{fig:bowtie_samples}
\end{figure}

To evaluate the efficacy of existing Unsupervised Anomaly Detection (UAD)
architectures on complex industrial topologies, we conducted our experiments on
the ``BowTie''~\cite{oceans11}
dataset. The raw acquisition set contains 1,360
labeled images across 10 trays (1,241 good and 119 reject). Unlike standard
benchmarks (e.g., MVTec AD \cite{bergmann2021mvtec}), this dataset presents
severe real-world manufacturing challenges, including highly reflective
metallic surfaces, microscopic structural defects, and background clutter.
Color profile assignments and counts are summarized in Table~\ref{tab:tray_profile_counts}, and representative images from these distinct color profiles are illustrated in Figure~\ref{fig:bowtie_samples}. Reported CP1-CP3 and Combined experiments exclude tray 27952G, using 1,224 images (1,130 nominal and 94 defect samples).

To align with UAD terminology in the remainder of the paper, we refer to ``good'' images as nominal and ``reject'' images as defect samples.

\begin{table}[!t]
    \centering
    \caption{Label counts and profile assignments for the raw acquisition set.}
    \label{tab:tray_profile_counts}
    \begin{tabular}{lccccc}
        \toprule
        \textbf{Tray}   & \textbf{Profile} & \textbf{Good} & \textbf{Reject} & \textbf{Subtotal} & \textbf{Used} \\
        \midrule
        116             & CP2              & 126           & 10              & 136               & Yes           \\
        117             & CP3              & 128           & 8               & 136               & Yes           \\
        118             & CP3              & 129           & 7               & 136               & Yes           \\
        119             & CP3              & 110           & 26              & 136               & Yes           \\
        120             & CP1              & 133           & 3               & 136               & Yes           \\
        121             & CP1              & 118           & 18              & 136               & Yes           \\
        27952F          & CP2              & 132           & 4               & 136               & Yes           \\
        27952G          &                  & 111           & 25              & 136               & No            \\
        27952H          & CP2              & 130           & 6               & 136               & Yes           \\
        27952J          & CP1              & 124           & 12              & 136               & Yes           \\
        \midrule
        \textbf{Totals} &                  & \textbf{1241} & \textbf{119}    & \textbf{1360}     &               \\
        \bottomrule
    \end{tabular}
\end{table}

BowTie is therefore not intended as a replacement for public benchmarks, but as
a deployment-oriented stress test in the sense argued for by recent IAD surveys
and benchmark papers.
Recent surveys \cite{Liu_2024} note that real manufacturing data often
combine subtle defect structure, scarce anomalies, clutter, and domain-specific
acquisition effects, while benchmark analyses \cite{xie2024imiadindustrialimageanomaly} show that model rankings can shift materially once evaluation settings move closer to industrial practice and
uniform cross-family comparisons are enforced. We use BowTie in that
spirit: to test whether Anomalib-accessible models remain robust under
reflective metallic surfaces, profile-specific appearance shifts, and limited
reject support.

To characterize reject-class heterogeneity, Table~\ref{tab:defect_type_profile_counts} summarizes defect-type counts
across CP1-CP3. While the dataset contains several anomaly types,
representative examples of three common defects are shown in Figure~\ref{fig:all_defect_examples}. Gouges
(a, b) exhibit an elongated, scratch-like morphology that often blends with
specular reflections. Inclusions (c, d, e) appear as isolated surface
particles with highly variable size and contrast across color profiles.
Finally, nicks (f, g) range from subtle to large edge defects. Ultimately,
this wide morphological diversity complicates precise localization and robust
anomaly detection.

\begin{table}[htbp]
    \centering
    \caption{Defect counts for reject images across CP1-CP3.}
    \label{tab:defect_type_profile_counts}
    \begin{tabular}{lccc}
        \toprule
        \textbf{Defect Type} & \textbf{CP1} & \textbf{CP2} & \textbf{CP3} \\
        \midrule
        gouge                & 11           & 0            & 5            \\
        inclusion            & 17           & 7            & 15           \\
        nick                 & 4            & 0            & 3            \\
        chip                 & 0            & 13           & 0            \\
        void                 & 0            & 0            & 17           \\
        inclusion/gouge      & 0            & 0            & 1            \\
        inclusion/gold       & 1            & 0            & 0            \\
        \midrule
        \textbf{Totals}      & \textbf{33}  & \textbf{20}  & \textbf{41}  \\
        \bottomrule
    \end{tabular}
\end{table}
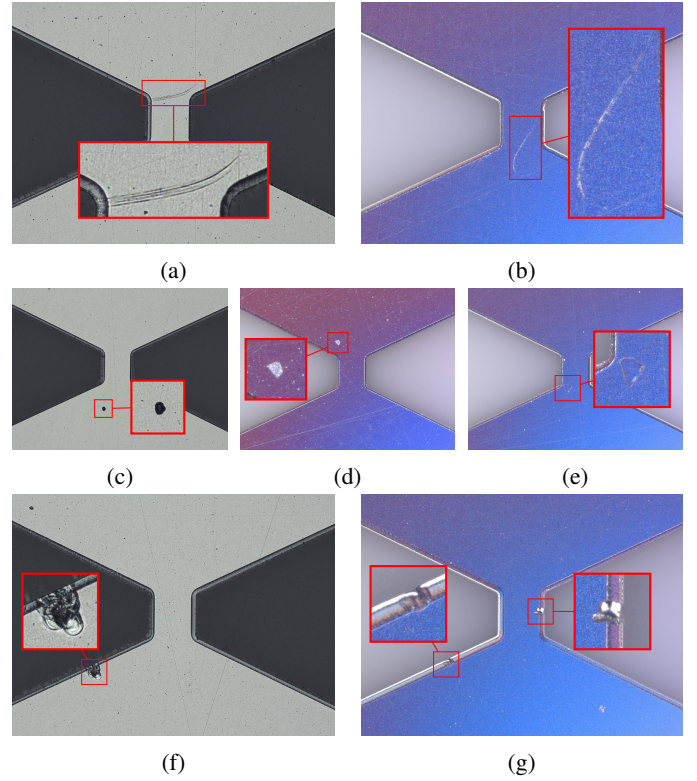
\begin{figure}[!t]
    \centering

    \begin{subfigure}[t]{0.48\linewidth}
        \centering
        \begin{tikzpicture}[
                spy using outlines={rectangle, magnification=3, width=2.5cm, height=1cm, connect spies}
            ]
            \node[inner sep=0pt, outer sep=0pt] (myimage)
                {\includegraphics[width=\textwidth]{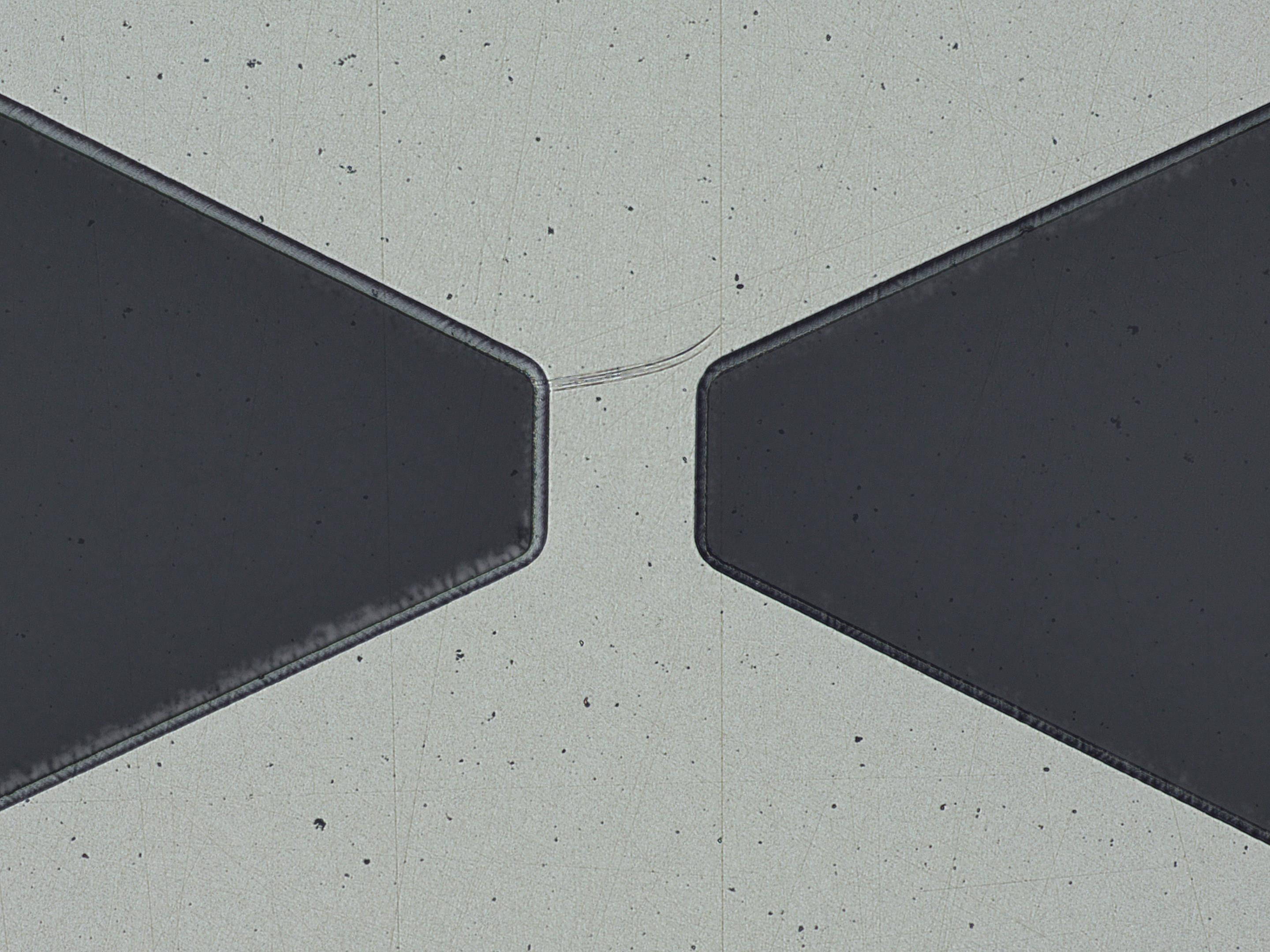}};
            \path[use as bounding box] (myimage.south west) rectangle (myimage.north east);

            \spy[red] on (0, 0.4) in node at (0, -0.75);
        \end{tikzpicture}
        \caption{}
        \label{fig:CP1_gouge}
    \end{subfigure}
    \hfill
    \begin{subfigure}[t]{0.48\linewidth}
        \centering
        \begin{tikzpicture}[
                spy using outlines={rectangle, magnification=3, width=1.25cm, height=2.5cm, connect spies}
            ]
            \node[inner sep=0pt, outer sep=0pt] (myimage)
                {\includegraphics[width=\textwidth]{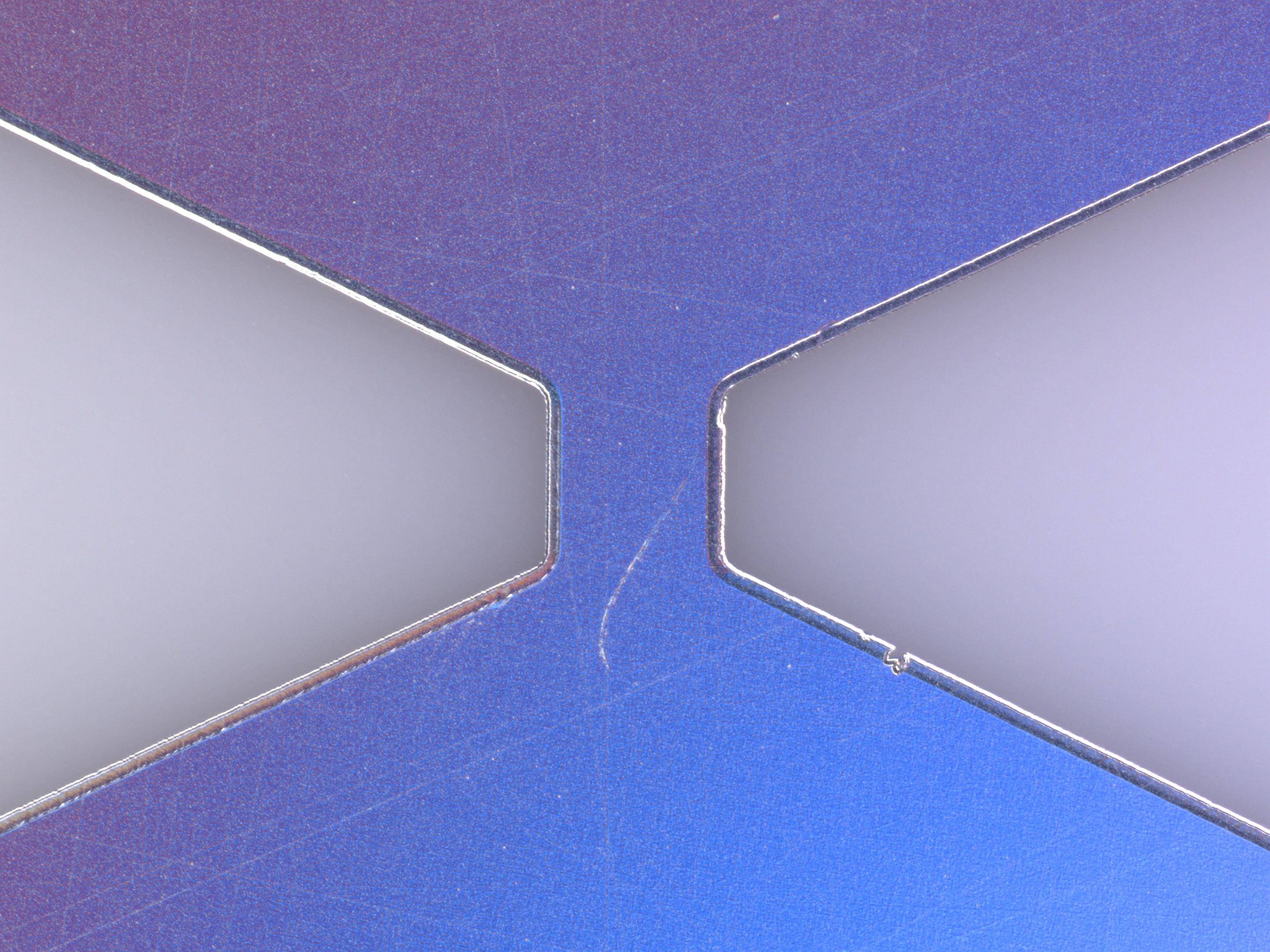}};
            \path[use as bounding box] (myimage.south west) rectangle (myimage.north east);

            \spy[red] on (0.05, -0.325) in node at (1.25, 0);
        \end{tikzpicture}
        \caption{}
        \label{fig:CP3_gouge}
    \end{subfigure}

    \vspace{.1em}

    \begin{subfigure}[t]{0.32\linewidth}
        \centering
        \begin{tikzpicture}[
                spy using outlines={rectangle, magnification=3, width=0.7cm, height=0.7cm, connect spies}
            ]
            \node[inner sep=0pt, outer sep=0pt] (myimage)
                {\includegraphics[width=\textwidth]{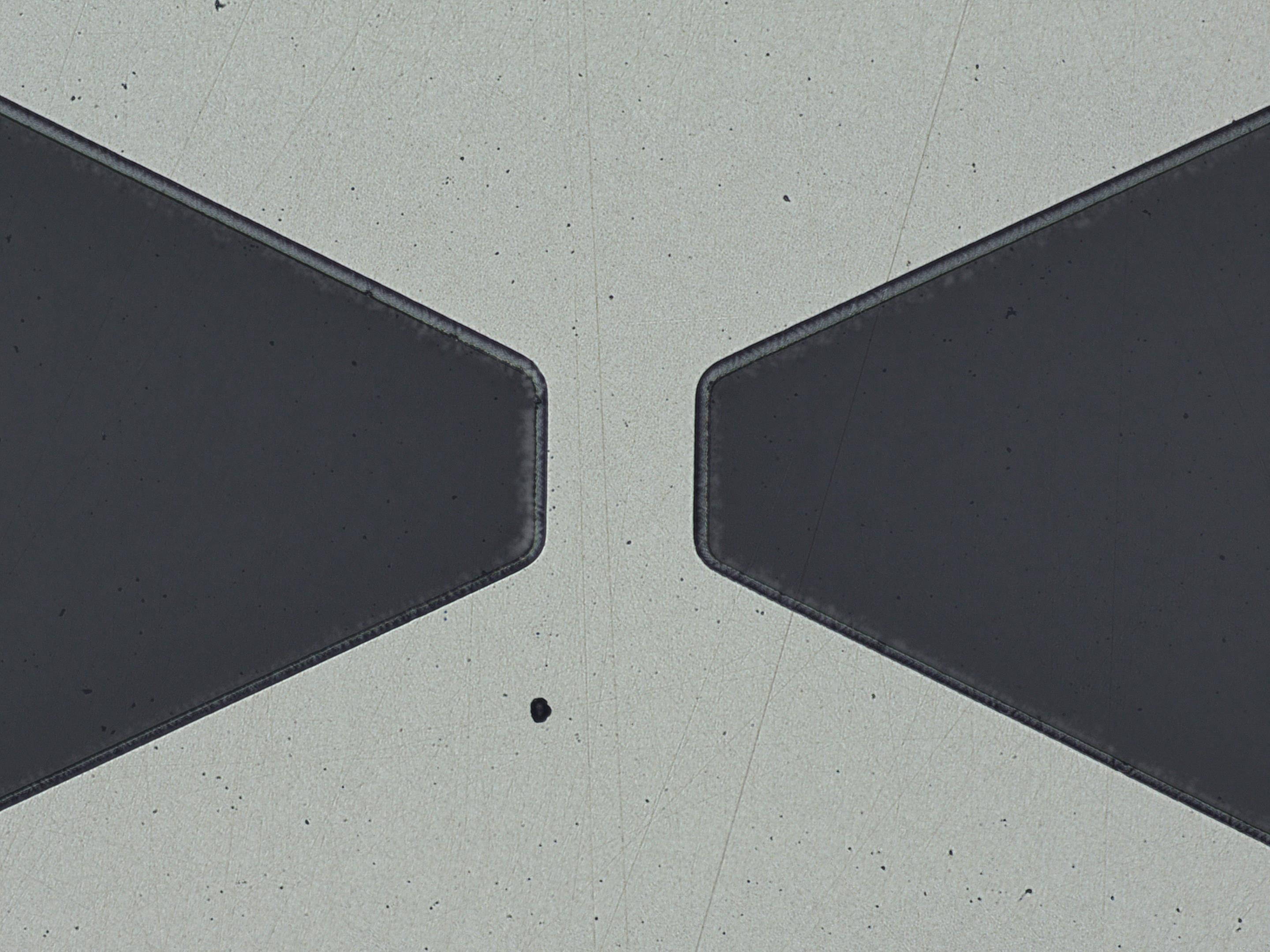}};
            \path[use as bounding box] (myimage.south west) rectangle (myimage.north east);

            \spy[red] on (-0.22, -0.52) in node at (0.5, -0.5);
        \end{tikzpicture}
        \caption{}
        \label{fig:CP1_inclusion}
    \end{subfigure}
    \hfill
    \begin{subfigure}[t]{0.32\linewidth}
        \centering
        \begin{tikzpicture}[
                spy using outlines={rectangle, magnification=3, width=0.8cm, height=0.8cm, connect spies}
            ]
            \node[inner sep=0pt, outer sep=0pt] (myimage)
                {\includegraphics[width=\textwidth]{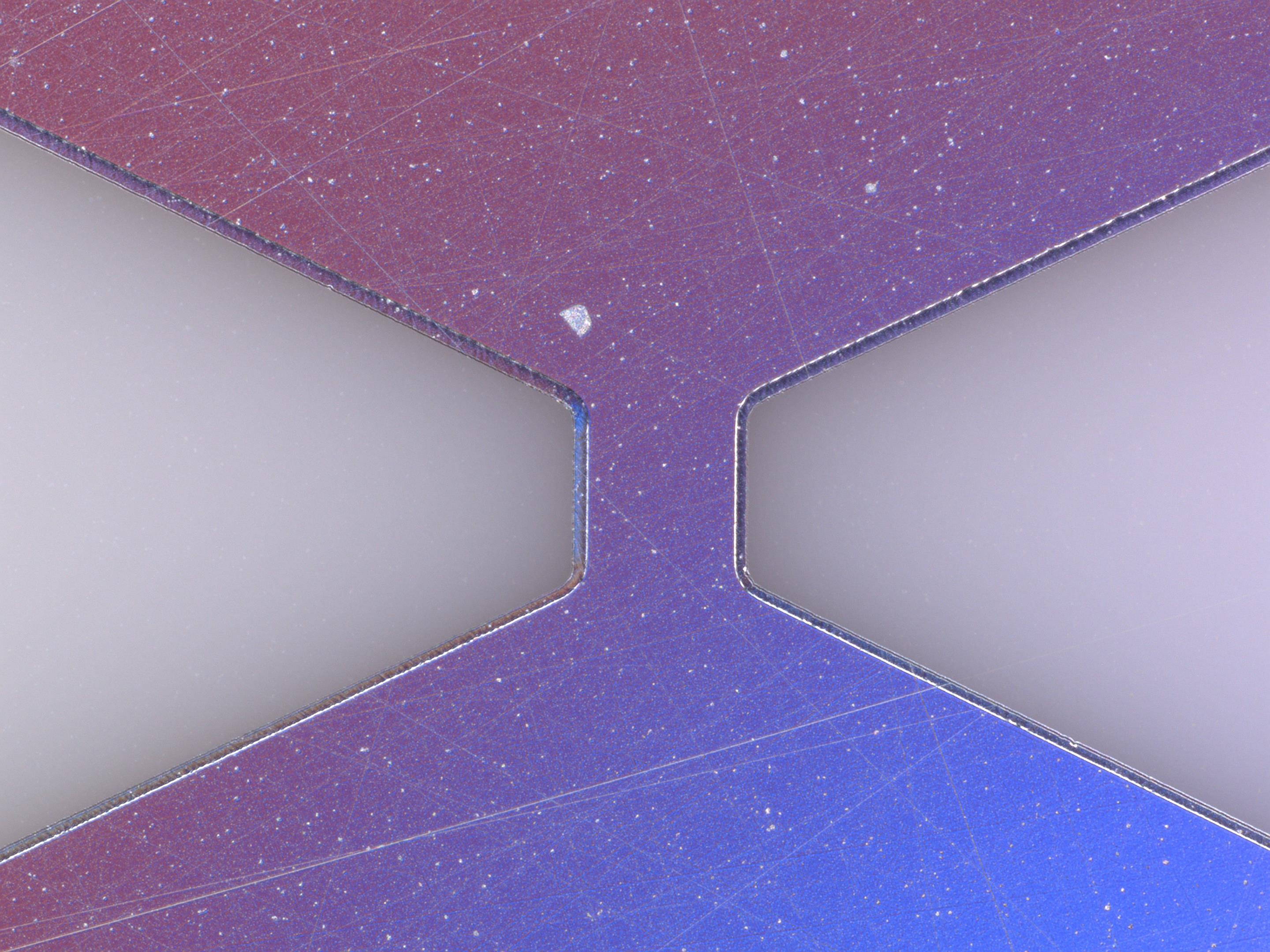}};
            \path[use as bounding box] (myimage.south west) rectangle (myimage.north east);

            \spy[red] on (-0.13, 0.35) in node at (-0.95, 0);
        \end{tikzpicture}
        \caption{}
        \label{fig:CP2_inclusion}
    \end{subfigure}
    \hfill
    \begin{subfigure}[t]{0.32\linewidth}
        \centering
        \begin{tikzpicture}[
                spy using outlines={rectangle, magnification=3, width=1cm, height=1cm, connect spies}
            ]
            \node[inner sep=0pt, outer sep=0pt] (myimage)
                {\includegraphics[width=\textwidth]{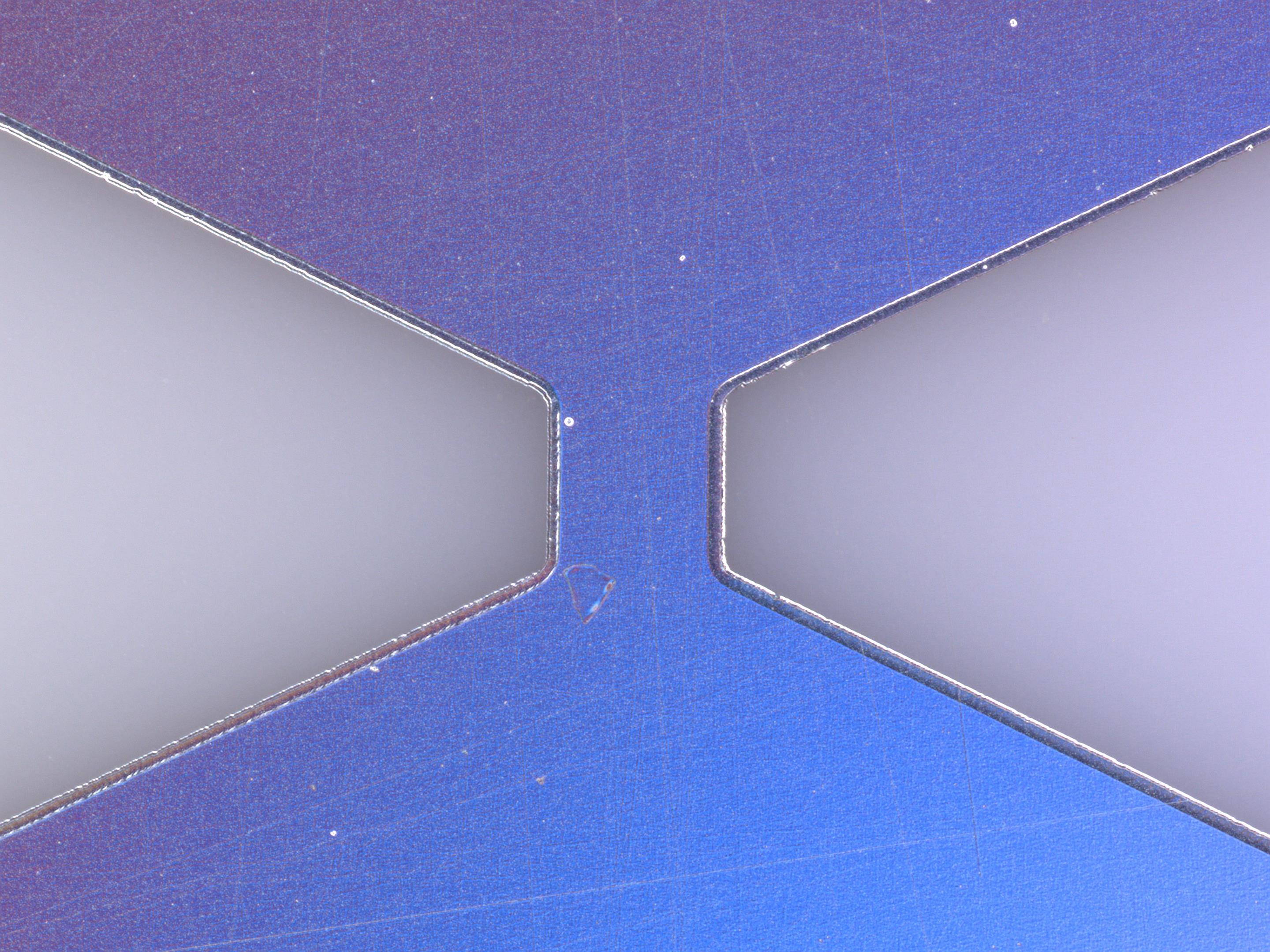}};
            \path[use as bounding box] (myimage.south west) rectangle (myimage.north east);

            \spy[red] on (-0.1, -0.25) in node at (.75, 0);
        \end{tikzpicture}
        \caption{}
        \label{fig:CP3_inclusion}
    \end{subfigure}

    \vspace{.1em}

    \begin{subfigure}[t]{0.48\linewidth}
        \centering
        \begin{tikzpicture}[
                spy using outlines={rectangle, magnification=3, width=1cm, height=1cm, connect spies}
            ]
            \node[inner sep=0pt, outer sep=0pt] (myimage)
                {\includegraphics[width=\textwidth]{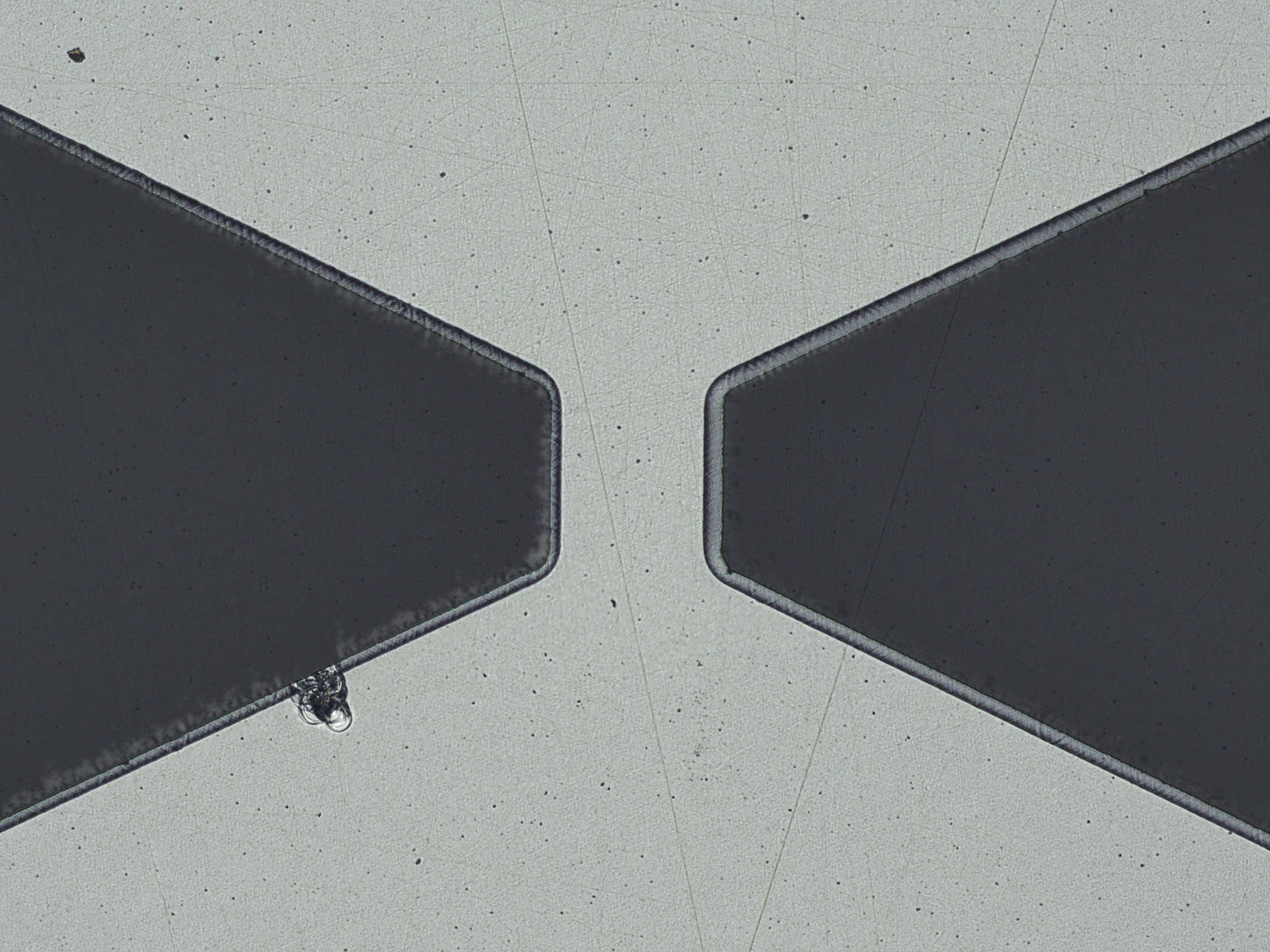}};
            \path[use as bounding box] (myimage.south west) rectangle (myimage.north east);

            \spy[red] on (-1.05, -0.75) in node at (-1.5, 0.05);
        \end{tikzpicture}
        \caption{}
        \label{fig:CP1_nick}
    \end{subfigure}
    \hfill
    \begin{subfigure}[t]{0.48\linewidth}
        \centering
        \begin{tikzpicture}[
                spy using outlines={rectangle, magnification=3, width=1cm, height=1cm, connect spies}
            ]
            \node[inner sep=0pt, outer sep=0pt] (myimage)
                {\includegraphics[width=\textwidth]{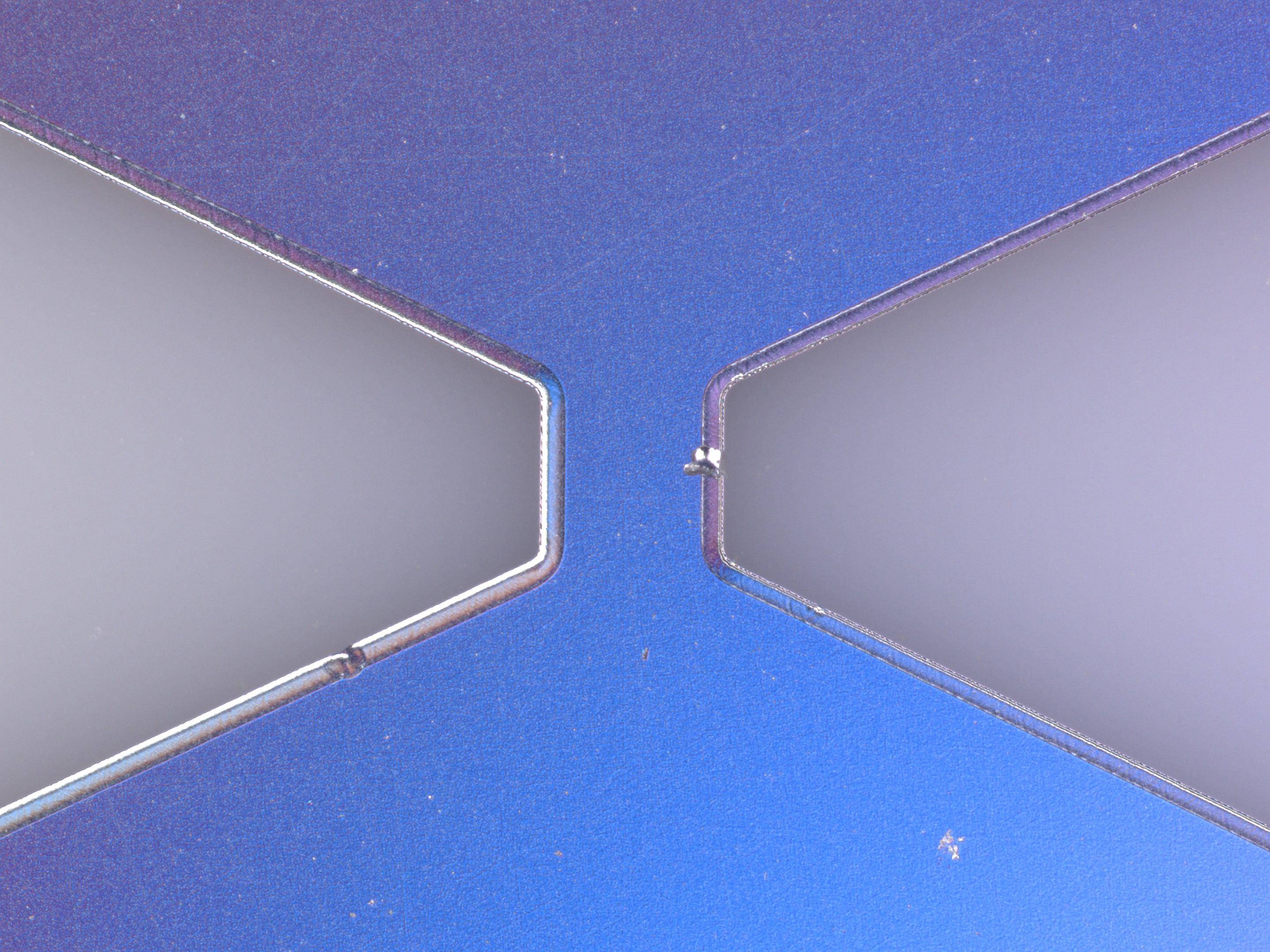}};
            \path[use as bounding box] (myimage.south west) rectangle (myimage.north east);

            \spy[red] on (0.25, 0.05) in node at (1.2, 0.05);
            \spy[red] on (-1, -0.65) in node at (-1.5, 0.15);
        \end{tikzpicture}
        \caption{}
        \label{fig:CP3_nick}
    \end{subfigure}

    \caption{Examples of reject-class defects across CP1-CP3. Red boxes indicate defect regions and magnified insets. Examples include gouges (a, b), inclusions (c, d, e), and nicks (f, g).}
    \label{fig:all_defect_examples}
\end{figure}

For model benchmarking, we split each color-profile pool at the image level and
train on nominal-only subsets.
20\% of the nominal pool
in each profile was withheld from training and then divided between validation
and test, while defect images were excluded from training and split between
validation and test within the same profile. Table~\ref{tab:profile_split_counts} gives the exact baseline train/validation/test
counts used in Section~\ref{sec:results}.

\begin{table}[!t]
    \centering
    \caption{Profile-wise baseline split counts. N = nominal and D = defect. Training uses nominal images only; validation and test contain held-out nominal images plus defect images from the same profile.}
    \label{tab:profile_split_counts}
    \scriptsize
    \setlength{\tabcolsep}{4pt}
    \begin{tabular}{lccccccc}
        \toprule
        \textbf{Profile} & \textbf{Train N} & \textbf{Val N} & \textbf{Val D} & \textbf{Val Tot.} & \textbf{Test N} & \textbf{Test D} & \textbf{Test Tot.} \\
        \midrule
        CP1              & 300              & 37             & 16             & 53                & 38              & 17              & 55                 \\
        CP2              & 311              & 38             & 10             & 48                & 39              & 10              & 49                 \\
        CP3              & 294              & 36             & 20             & 56                & 37              & 21              & 58                 \\
        \bottomrule
    \end{tabular}
    \vspace{-5mm}
\end{table}
\subsection{Evaluated Architectures}
\label{subsec:architectures}
We evaluated 19 Anomalib-compatible UAD models \cite{akcay2022anomalib} spanning five recurring families: student-teacher discrepancy methods; patch-memory / feature-adaptation methods; flow-based density estimators; reconstruction-based methods; and feature-space density baselines \cite{batzner2024efficientadaccuratevisualanomaly,deng2022anomalydetectionreversedistillation,wang2021studentteacherfeaturepyramidmatching,
    roth2022totalrecallindustrialanomaly,
    defard2020padimpatchdistributionmodeling,lee2022cfacoupledhyperspherebasedfeatureadaptation,damm2025anomalydinoboostingpatchbasedfewshot,rolih2024supersimplenetunifyingunsupervisedsupervised, yu2021fastflowunsupervisedanomalydetection,gudovskiy2021cflowadrealtimeunsupervisedanomaly,rudolph2021fullyconvolutionalcrossscaleflowsimagebased,tailanian2024uflowushapednormalizingflow, guo2025dinomalyphilosophymulticlassunsupervised,zavrtanik2021draemdiscriminativelytrained,zavrtanik2022dsrdualsubspace,akcay2018ganomalysemisupervisedanomalydetection,ndiour2022frefastmethodanomaly,ahuja2019probabilisticmodelingdeepfeatures}.
Several entries are hybrids rather than pure exemplars, so these family labels are used as interpretive shorthand rather than as claims that each method instantiates a single mechanism.

This family-oriented framing follows the broader IAD literature, which frequently organizes unsupervised methods into recurring mechanism classes and treats family-level comparisons as more informative than isolated per-model rankings \cite{Liu_2024,xie2024imiadindustrialimageanomaly}. Keeping BowTie aligned with that taxonomy makes it easier to distinguish broad mechanism behavior from quirks of a single implementation.

\subsection{Implementation Details}
\label{subsec:implementation}
All experiments utilized the Anomalib framework \cite{akcay2022anomalib}. To keep the comparison fair without negating the assumptions made in the original papers, we standardized the data splits, reporting metrics, and early-stopping policy, but we did not force all methods into a single surrogate anomaly score. Student-teacher models retained feature-discrepancy scoring; patch-based models retained nearest-neighbor or patch-density scoring; flow-based methods retained likelihood estimation; and reconstruction-driven methods retained reconstruction or reconstruction-error outputs.

Most architectures were trained with a baseline input size of $256 \times 256$, although a small number of configurations retained a higher configured baseline resolution. Specifically, U-Flow and Dinomaly retain a higher configured baseline of $448 \times 448$ due to Anomalib implementation constraints. The grayscale and resolution settings should therefore be read as controlled sensitivity probes around each model's configured sweep setting rather than as an attempt to collapse all architectures into an identical preprocessing regime.

To ensure optimal convergence without overfitting, each configuration
used early stopping with a patience of 20 epochs. BowTie does not
include pixel-level ground-truth masks, so model selection and
threshold selection were both performed at the image level. F1-Max
denotes the highest F1 score achievable by sweeping all possible
decision thresholds; it is used solely to identify the optimal
threshold on the validation split. Macro F1 is then computed at
that fixed threshold on the test split, averaging equally over the
nominal and anomalous classes to account for the inherent class
imbalance between nominal and defective samples. Each model,
profile, and preprocessing setting used the seeded split above.

We provide code for reproducibility
\cite{experiments-reproducibility}.

\subsection{Consensus-Based Data Audit}
\label{subsec:sanitization}
After running the full benchmark across the available architectures and the preprocessing ablations discussed in Section~\ref{sec:results}, we observed that performance repeatedly plateaued below the behavior typically expected on cleaner anomaly-detection benchmarks. That pattern suggested that model choice alone was not explaining the remaining error and motivated a secondary audit of nominal-data quality. Because the benchmark pipeline already produced inference outputs for every model, we reused those outputs as a diagnostic signal rather than attempting unsupported manual relabeling.

Each model was therefore run on its nominal training split, producing image-level anomaly indications for samples that had been treated as normal during optimization. We then aggregated the model outputs and marked any training image flagged as anomalous by six or more benchmark models as a consensus-flagged nominal sample. This six-model threshold requires cross-family agreement while keeping the audit conservative for SME triage. This consensus rule is used solely to prioritize targeted SME review, not to create new labels. It applies to training-set predictions only and does not reference test labels. Because the benchmark includes memory banks, one-class embeddings, flow likelihood models, and reconstruction-driven detectors, repeated agreement across families serves here as a practical review signal rather than as a surrogate ground-truth label. Reruns that exclude this consensus-flagged nominal subset from the training pool are interpreted in Section~\ref{sec:label_noise} as a discussion-oriented data-quality analysis rather than as part of the primary benchmark results.

\section{Results}
\label{sec:results}

This section moves from raw scores to operational interpretation. Table~\ref{tab:baseline_taxonomy_matrix} summarizes which architectural families remain reliable on the BowTie dataset at baseline. The consensus-based contamination audit described in Section~\ref{subsec:sanitization} is interpreted in Section~\ref{sec:label_noise} as a secondary discussion of data quality rather than a primary benchmark result.

\subsection{Architectural Robustness to Complex Industrial Data}
\begin{figure*}[!t]
    \centering
    \includegraphics[width=0.49\textwidth]{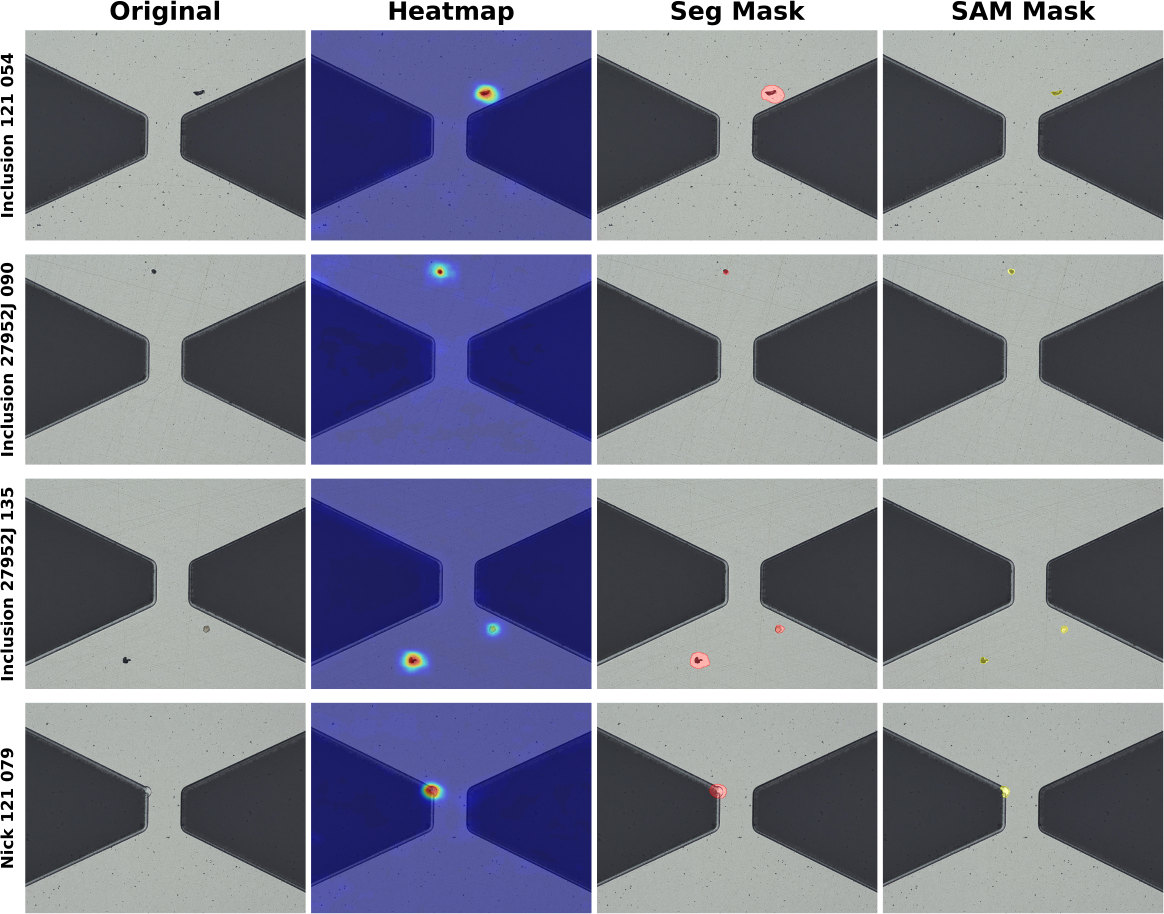}
    \hfill
    \includegraphics[width=0.49\textwidth]{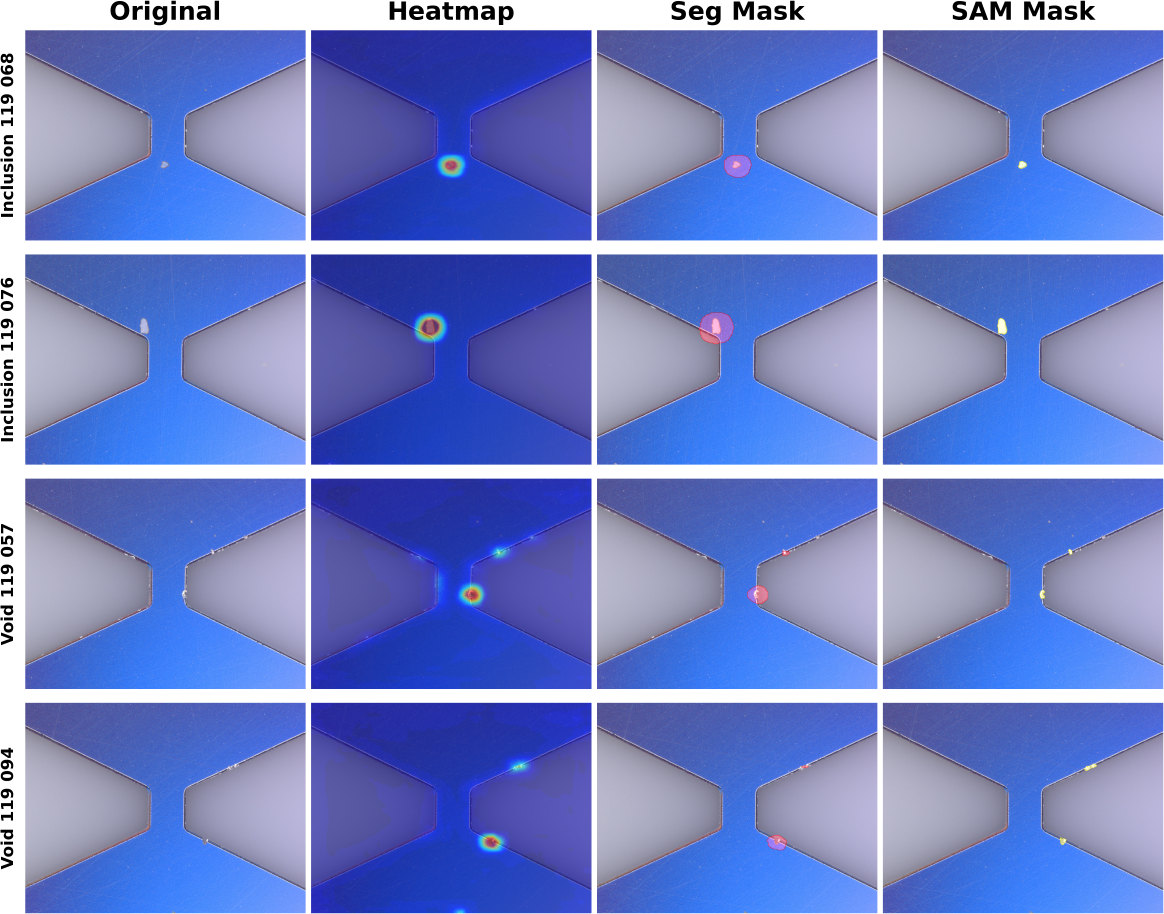}
    \caption{EfficientAD detection results on four CP1 (left) and four CP3 (right) defect samples: originals, heatmaps, overlaid masks, and SAM-refined proposals \cite{ravi2024sam2}. In the workflow, heatmaps locate candidate defect vicinity and SAM sharpens regions for inspector acceptance, rejection, or boundary adjustment; these masks are not benchmark ground truth.}
    \label{fig:cp1-cp3-matrices}
    \vspace{-3mm}
\end{figure*}

BowTie is a challenging benchmark for unsupervised anomaly detection because reflective metallic surfaces, microscopic scratches, and cluttered backgrounds violate the cleaner appearance assumptions that underlie standard benchmarks such as MVTec AD \cite{bergmann2021mvtec}. Table~\ref{tab:baseline_taxonomy_matrix} therefore serves as the main baseline summary for the cross-profile comparison, while Figure~\ref{fig:cp1-cp3-matrices} is used afterward as a qualitative check on where those high-ranking models place their attention.

Prior work shows that nominal-data quality, imaging variation, and deployment constraints can reorder methods that appear stable on cleaner benchmarks \cite{Liu_2024,xie2024imiadindustrialimageanomaly}. BowTie extends that perspective to a reflective metal-inspection setting with explicit cross-profile appearance shift.

\begin{table}[!t]
    \centering
    \caption{Baseline performance across color profiles, grouped by architectural taxonomy. Best values bold; second-best values, including ties, underlined. M-F1 = Macro F1, A-F1 = Anomalous F1. The Combined column reports results from a single training run pooling all three color-profile subsets into one dataset.}
    \label{tab:baseline_taxonomy_matrix}
    \setlength{\tabcolsep}{3pt}
    \renewcommand{\arraystretch}{1.0}
    \scriptsize
    \resizebox{\linewidth}{!}{%
        \begin{tabular}{@{}lcccccccc@{}}
            \toprule
            \textbf{Model} & \multicolumn{2}{c}{\textbf{CP1}} & \multicolumn{2}{c}{\textbf{CP2}} & \multicolumn{2}{c}{\textbf{CP3}} & \multicolumn{2}{c}{\textbf{Combined}}                                                                         \\
            \cmidrule(lr){2-3} \cmidrule(lr){4-5} \cmidrule(lr){6-7} \cmidrule(lr){8-9}
                           & \textbf{M-F1}                    & \textbf{A-F1}                    & \textbf{M-F1}                    & \textbf{A-F1}                      & \textbf{M-F1}   & \textbf{A-F1}   & \textbf{M-F1}   & \textbf{A-F1}   \\
            \midrule
            \multicolumn{9}{@{}l}{\textit{Knowledge Distillation}}                                                                                                                                                                               \\
            EfficientAD    & \best{.872}                      & \best{.824}                      & \best{.760}                      & \best{.667}                        & .716            & .667            & .740            & .608            \\
            STFPM          & .742                             & .667                             & .486                             & .444                               & .680            & .625            & .721            & .612            \\
            Reverse Dist.  & .811                             & .743                             & .578                             & .500                               & \runnerup{.797} & \runnerup{.744} & .715            & .624            \\
            \multicolumn{9}{@{}l}{\textit{Memory Bank / Feature Adaptation}}                                                                                                                                                                     \\
            PatchCore      & .817                             & .757                             & .500                             & .429                               & \best{.820}     & \best{.783}     & .720            & .650            \\
            AnomalyDINO    & .735                             & .649                             & .493                             & .400                               & .612            & .512            & .655            & .481            \\
            PaDiM          & .663                             & .480                             & \runnerup{.738}                  & \runnerup{.615}                    & .767            & .667            & .619            & .550            \\
            SuperSimpleNet & .675                             & .605                             & .667                             & .476                               & .623            & .429            & .569            & .404            \\
            CFA            & .686                             & .571                             & .623                             & .400                               & .566            & .603            & .565            & .454            \\
            \multicolumn{9}{@{}l}{\textit{Normalizing Flows}}                                                                                                                                                                                    \\
            CFlow          & \runnerup{.863}                  & .800                             & .170                             & .339                               & .721            & .692            & .708            & .575            \\
            CS-Flow        & .692                             & .619                             & .682                             & .552                               & .703            & .667            & .706            & .620            \\
            U-Flow         & .760                             & .643                             & .686                             & .538                               & \runnerup{.797} & \runnerup{.744} & .690            & .595            \\
            FastFlow       & .861                             & \runnerup{.821}                  & .566                             & .424                               & .635            & .667            & .631            & .495            \\
            \multicolumn{9}{@{}l}{\textit{Reconstruction / GANs}}                                                                                                                                                                                \\
            Dinomaly       & .861                             & \runnerup{.821}                  & .712                             & .600                               & .744            & .682            & \best{.771}     & \best{.680}     \\
            DRAEM          & .719                             & .606                             & .657                             & .545                               & .686            & .605            & \runnerup{.755} & \runnerup{.667} \\
            FRE            & .592                             & .488                             & .650                             & .500                               & .653            & .677            & .547            & .462            \\
            GANomaly       & .236                             & .472                             & .435                             & .100                               & .295            & .538            & .442            & .473            \\
            DSR            & .647                             & .596                             & .477                             & .125                               & .437            & .563            & .388            & .472            \\
            \multicolumn{9}{@{}l}{\textit{Feature Statistics / Density Estimation}}                                                                                                                                                              \\
            DFM            & .613                             & .486                             & .612                             & .485                               & .735            & .694            & .595            & .395            \\
            DFKDE          & .293                             & .486                             & .592                             & .333                               & .613            & .560            & .243            & .451            \\
            \bottomrule
        \end{tabular}%
    }
    \vspace{-4mm}
\end{table}
At baseline, several student-teacher discrepancy models and patch-level nominal-support methods appear near the top of the benchmark. EfficientAD leads CP1 and CP2, PatchCore leads CP3, and Dinomaly is strongest on the Combined baseline. These baseline rankings indicate that strong BowTie performance is achievable in multiple families, but they do not isolate a single reason for that performance.

Other families are more profile-dependent. Flow models vary sharply:
CFlow is excellent on CP1 but weak on CP2, whereas CS-Flow and U-Flow
are steadier. Reconstruction-driven methods are similarly mixed, with
Dinomaly strong and DRAEM, DSR, and GANomaly less stable across
profiles. These patterns suggest that model-specific behavior may
matter at least as much as family labels, so the groupings are
descriptive rather than mechanistic.

The anomaly maps in Figure~\ref{fig:cp1-cp3-matrices} complement Table~\ref{tab:baseline_taxonomy_matrix} by showing whether the strongest models actually localize defect-like structure or merely react to specular edges and benign texture shifts. In the intended workflow, the heatmap provides coarse defect localization and the accompanying SAM masks illustrate how those regions can be refined into inspector-facing candidate segments for acceptance, rejection, or boundary adjustment; they are not benchmark ground truth.

\subsection{The Brittleness of Representations}

The grayscale and resolution sweeps show that representation stability
is architecture-dependent rather than family-uniform. The flow
papers in this benchmark all model likelihoods in feature space,
but BowTie shows that their sensitivity differs sharply across
implementations:
CFlow swings from a major loss on CP1 to a major gain on CP2, whereas
CS-Flow and U-Flow are much less erratic, as shown in Table~\ref{tab:sensitivity_matrix}.

\begin{table}[h]
    \centering
    \caption{Sensitivity matrix for grayscale (G) and resolution scaling ($350{\times}350$, $448{\times}448$) on the CP1 top-10 models from Table~\ref{tab:cleaning_matrix}, ranked by CP1 post-audit Macro F1. Cells show signed $\Delta$ Macro F1 vs.\ baseline. $^\dagger$Models already at $448{\times}448$ in baseline. Dashes: ablation omitted due to Anomalib limits.}
    \label{tab:sensitivity_matrix}
    \setlength{\tabcolsep}{4pt}
    \renewcommand{\arraystretch}{1.15}
    \footnotesize
    \resizebox{\linewidth}{!}{%
        \begin{tabular}{@{}lccccccccc@{}}
            \toprule
            \textbf{Model}     & \multicolumn{3}{c}{\textbf{CP1}} & \multicolumn{3}{c}{\textbf{CP2}} & \multicolumn{3}{c}{\textbf{CP3}}                                                                                       \\
            \cmidrule(lr){2-4} \cmidrule(lr){5-7} \cmidrule(lr){8-10}
                               & \textbf{G}                       & \textbf{350}                     & \textbf{448}                     & \textbf{G} & \textbf{350} & \textbf{448} & \textbf{G} & \textbf{350} & \textbf{448} \\
            \midrule
            Dinomaly$^\dagger$ & $-$.040                          & $\pm$.000                        & $\pm$.000                        & $+$.026    & $+$.054      & $+$.012      & $-$.028    & $+$.043      & $+$.021      \\
            DRAEM              & $+$.009                          & --                               & $-$.127                          & $-$.052    & --           & $-$.016      & $-$.283    & --           & $+$.026      \\
            EfficientAD        & $-$.033                          & $+$.004                          & $-$.096                          & $-$.016    & $+$.052      & $+$.010      & $+$.017    & $-$.006      & $+$.060      \\
            STFPM              & $-$.031                          & $+$.034                          & $+$.039                          & $-$.098    & $+$.074      & $+$.067      & $-$.267    & $+$.107      & $+$.146      \\
            PatchCore          & $+$.032                          & $+$.018                          & $+$.046                          & $+$.133    & $+$.112      & $+$.271      & $-$.111    & $-$.101      & $+$.048      \\
            Reverse Dist.      & $+$.024                          & $-$.047                          & $+$.006                          & $+$.055    & $+$.246      & $+$.223      & $-$.064    & $+$.016      & $+$.012      \\
            CFlow              & $-$.440                          & $-$.250                          & $-$.413                          & $+$.558    & $+$.390      & $+$.315      & $+$.020    & $-$.112      & $-$.256      \\
            CS-Flow            & $+$.033                          & --                               & --                               & $+$.030    & --           & --           & $-$.101    & --           & --           \\
            U-Flow$^\dagger$   & $+$.021                          & $+$.116                          & $+$.030                          & $+$.019    & $+$.085      & $+$.038      & $-$.062    & $-$.005      & $-$.021      \\
            AnomalyDINO        & $-$.049                          & $-$.076                          & $\pm$.000                        & $+$.130    & $+$.235      & $+$.226      & $-$.110    & $-$.288      & $+$.026      \\
            \bottomrule
        \end{tabular}%
    }
\end{table}
By contrast, PatchCore remains positive under grayscale on CP1 and
CP2, and Dinomaly changes only moderately across all three profiles.
Those patterns are consistent with patch-memory matching and
Transformer-feature representations retaining useful signal when
chromatic cues are removed, although this study does not isolate
that cause.

Resolution exposes a different trade-off. PatchCore improves at $448 \times 448$ on all three profiles, and Reverse Distillation, STFPM, and AnomalyDINO also gain strongly on the harder profiles. The underlying papers all emphasize local patch comparison or multiscale feature matching, so these gains are compatible with their published design goals, although BowTie alone does not show whether the improvement comes specifically from finer defect detail, other scale effects, or interactions with the training protocol \cite{roth2022totalrecallindustrialanomaly,deng2022anomalydetectionreversedistillation,wang2021studentteacherfeaturepyramidmatching,damm2025anomalydinoboostingpatchbasedfewshot}. The flow models again do not move in unison. CFlow improves on CP2 but degrades on CP1 and CP3, whereas U-Flow stays closer to baseline. For industrial deployment, the results support treating grayscale conversion and sensor resolution as model-specific tuning variables rather than universal preprocessing defaults.

Taken together, the baseline table and reported ablations show that BowTie performance depends on more than a single leaderboard rank. EfficientAD, PatchCore, Dinomaly, Reverse Distillation, and CS-Flow appear among the strongest candidates in the reported experiments, based on baseline rank and observed robustness. This distinction between rank and stability motivates the data-quality discussion that follows.

\section{Discussion}
\label{sec:label_noise}

Consistent with recent IAD surveys and benchmark analyses \cite{Liu_2024,xie2024imiadindustrialimageanomaly}, BowTie shows a relatively consistent performance ceiling across several model families. That ceiling is informative because the evaluated models encode nominality in very different ways: memory banks store representative normal patches, PaDiM- and DFM-style baselines fit feature distributions, student-teacher methods compress one-class embeddings, and flow models estimate feature likelihoods \cite{roth2022totalrecallindustrialanomaly,defard2020padimpatchdistributionmodeling,ahuja2019probabilisticmodelingdeepfeatures,deng2022anomalydetectionreversedistillation,gudovskiy2021cflowadrealtimeunsupervisedanomaly,rudolph2021fullyconvolutionalcrossscaleflowsimagebased}. When that many mechanisms plateau on the same dataset, architecture alone may not fully explain the remaining error. Rather than asking non-experts to relabel marginal cases by hand, we used the model outputs already produced by the benchmark as a diagnostic signal. As defined in Section~\ref{subsec:sanitization}, the consensus pass therefore serves as a practical audit signal when SME review time is limited. This interpretation is also compatible with feature-distribution views of reliability, where repeatedly flagged nominal images can be interpreted as resembling samples near the edge of the learned normal manifold \cite{ahuja2019probabilisticmodelingdeepfeatures}.

\begin{table}[!t]
    \centering
    \caption{Post-audit comparison after removing the consensus-flagged nominal subset for the CP1 top-10 models, ranked by CP1 post-audit Macro F1. Cells show post-audit Macro F1 with relative \% change from baseline Macro F1 (green = improvement, red = degradation; saturation encodes magnitude). Best post-audit Macro F1 per profile is shown in bold.}
    \label{tab:cleaning_matrix}
    \setlength{\tabcolsep}{6pt}
    \renewcommand{\arraystretch}{1.15}
    \footnotesize
    \resizebox{\linewidth}{!}{%
        \begin{tabular}{@{}lccc@{}}
            \toprule
            \textbf{Model} & \textbf{CP1}        & \textbf{CP2}         & \textbf{CP3}        \\
            \midrule
            Dinomaly       & \cellpm{\best{.956}}{11.0} & \cellpw{.735}{3.2}   & \cellpm{\best{.849}}{14.1} \\
            U-Flow         & \cellps{.933}{22.8} & \cellnw{.682}{0.6}   & \cellnm{.723}{9.3}  \\
            EfficientAD    & \cellpw{.884}{1.4}  & \cellnw{.744}{2.1}   & \cellnw{.684}{4.5}  \\
            CFlow          & \cellnw{.852}{1.3}  & \cellps{.531}{212.4} & \cellns{.309}{57.1} \\
            Reverse Dist.  & \cellpw{.831}{2.5}  & \cellps{.779}{34.8}  & \cellnw{.782}{1.9}  \\
            PatchCore      & \cellnw{.798}{2.3}  & \cellps{.764}{52.8}  & \cellnm{.723}{11.8} \\
            AnomalyDINO    & \cellpm{.772}{5.0}  & \cellpm{.552}{12.0}  & \cellpm{.657}{7.4}  \\
            CS-Flow        & \cellpm{.748}{8.1}  & \cellpm{.757}{11.0}  & \cellpm{.808}{14.9} \\
            DRAEM          & \cellnw{.689}{4.2}  & \cellps{\best{.852}}{29.7}  & \cellns{.360}{47.5} \\
            STFPM          & \cellnm{.689}{7.1}  & \cellps{.592}{21.8}  & \cellnm{.596}{12.4} \\
            \bottomrule
        \end{tabular}
    }
    \vspace{-5mm}
\end{table}
Table~\ref{tab:cleaning_matrix} summarizes what happens when the consensus-flagged nominal subset identified by the audit is removed and the cleaned reruns are compared against baseline. The largest gains occur on CP2, the same profile that was most sensitive to grayscale and resolution choices. The most striking individual result is CFlow, which recovers from very low baseline performance (Macro F1 = 0.170) to 0.531 after cleaning---a result to interpret cautiously given the small CP2 test support, but one that illustrates why nominal-data auditing is especially consequential for likelihood-based architectures \cite{gudovskiy2021cflowadrealtimeunsupervisedanomaly}. That pattern is consistent with an interaction between profile-specific appearance variation and the consensus-flagged subset, but it does not by itself establish that these samples materially altered the learned support. The methods that benefit most, such as PatchCore, Reverse Distillation, CS-Flow, and Dinomaly, rely on comparatively tight nominal embeddings or likelihood models in their original formulations, so their improvements after cleaning are compatible with that interpretation without proving it \cite{roth2022totalrecallindustrialanomaly,deng2022anomalydetectionreversedistillation,rudolph2021fullyconvolutionalcrossscaleflowsimagebased,guo2025dinomalyphilosophymulticlassunsupervised}. Cleaning is not universally helpful, and that is part of the point: some architectures regress after removal of the flagged subset, especially on CP3, suggesting that flagged samples warrant selective review rather than blanket removal \cite{zavrtanik2021draemdiscriminativelytrained,zavrtanik2022dsrdualsubspace,akcay2018ganomalysemisupervisedanomalydetection,gudovskiy2021cflowadrealtimeunsupervisedanomaly}.

The combined results suggest two practical deployment considerations. First, preprocessing should be validated jointly with architecture rather than inherited as a default recipe, since grayscale and resolution effects are strongly method-dependent \cite{zhang2023makesgooddataaugmentation}. Second, nominal-data auditing should be treated as part of model qualification rather than as a one-time dataset cleanup step, especially when continuous SME review is not available \cite{ahuja2019probabilisticmodelingdeepfeatures}. The comparison should be read as a BowTie-specific stress test rather than as a claim that every model is operating in its ideal published setting \cite{damm2025anomalydinoboostingpatchbasedfewshot,guo2025dinomalyphilosophymulticlassunsupervised,batzner2024efficientadaccuratevisualanomaly,xie2024imiadindustrialimageanomaly}. Taken together, these findings motivate a workflow in which annotation, inference, and quantitative validation are tightly coupled rather than handled in isolation.

\section{Tool Overview}
\label{sec:tool_overview}

\begin{figure}[!t]
    \centering
    \includegraphics[width=\linewidth]{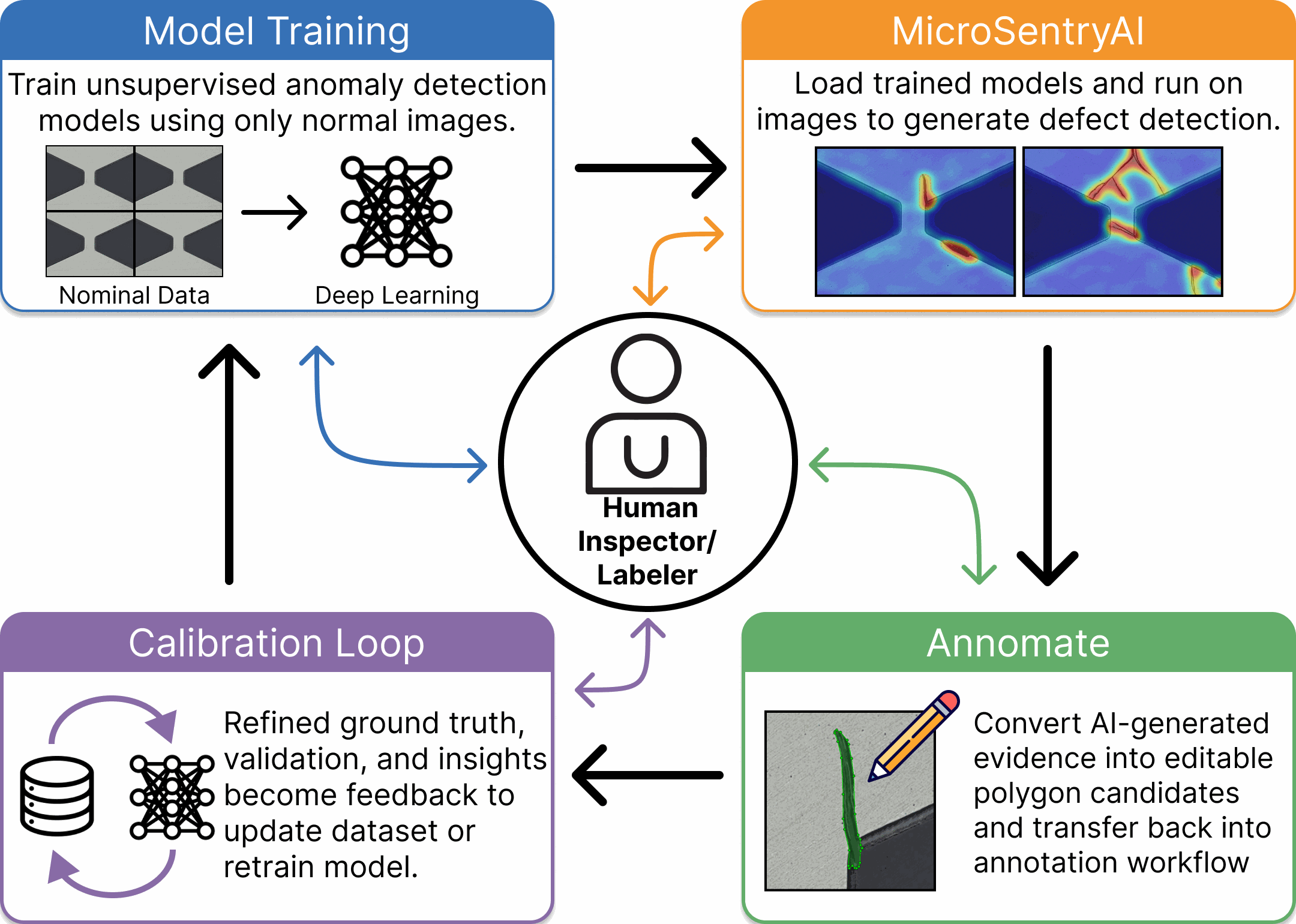}
    \caption{AnnoMate \& MicroSentryAI human-in-the-loop lifecycle}
    \label{fig:workflow_diagram}
    \vspace{-4mm}
\end{figure}

With the benchmark findings established, this section turns to the software suite initially deployed in the motivating inspection workflow. Figure~\ref{fig:workflow_diagram} shows annotation, inference, and validation in one graphical workflow while final accept/reject authority remains with the human inspector. At submission, the framework was in initial operational testing with inspectors; formal measures of inspection time, consistency, and onboarding effects remain future work.

\subsection{Architecture Summary}
The design emphasizes modular, human-in-the-loop inspection. \textit{AnnoMate}, \textit{MicroSentryAI}, and the validation module remain decoupled while exchanging annotations, CSV exports, and Anomalib-compatible outputs. Inspectors can review predictions, tune inference parameters, and return accepted AI-generated masks to annotation without leaving the interface.

\subsection{Workflow Summary}
\begin{figure}[!t]
    \centering
    \includegraphics[width=0.95\columnwidth]{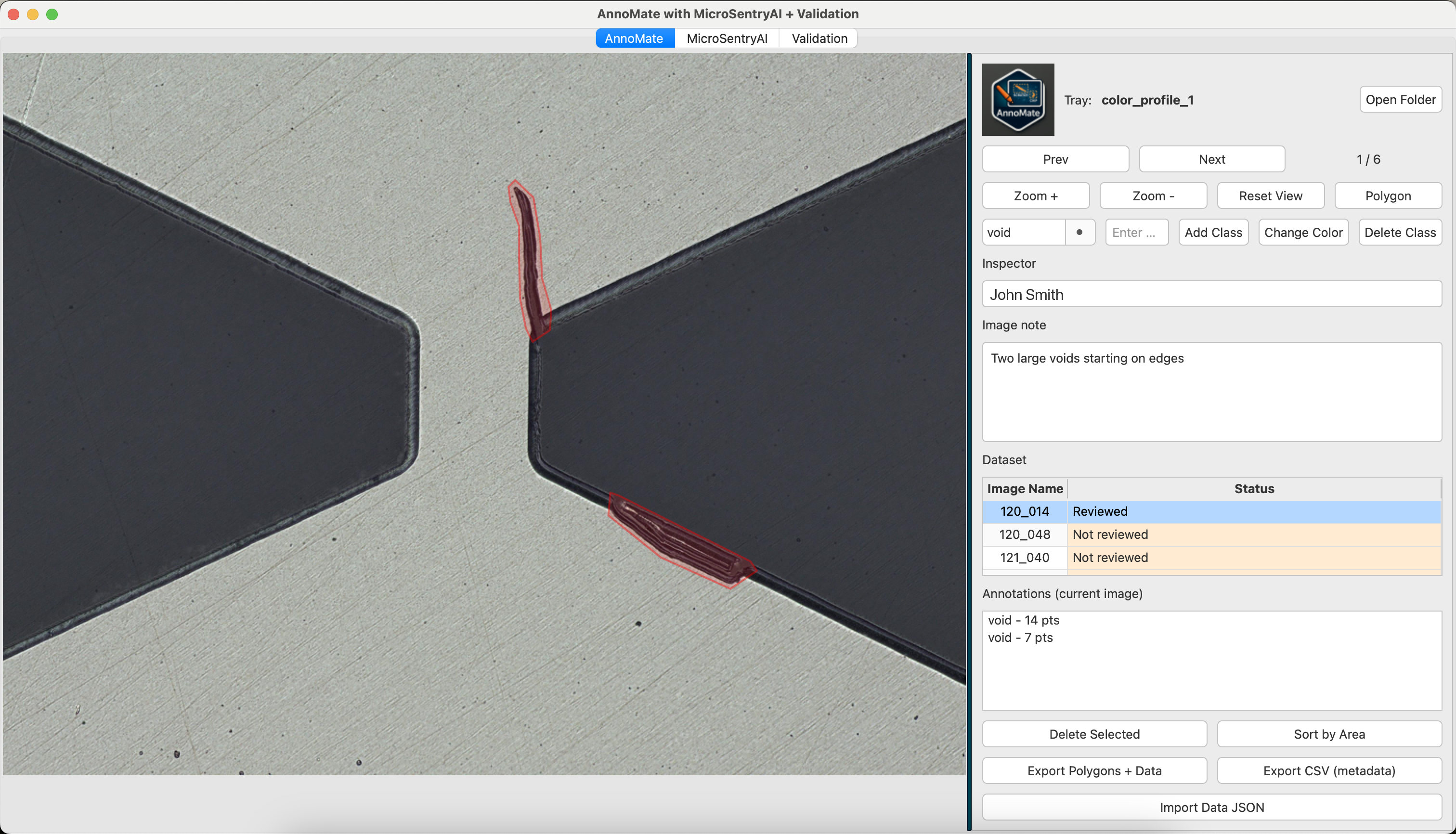}
    \caption{The \textit{AnnoMate} interface for defect annotation. It supports polygon-based labeling, metadata tracking, and dataset review to streamline ground-truth curation.}
    \label{fig:my_image_1}
    \vspace{-1mm}
\end{figure}

\textit{AnnoMate} serves as the annotation and curation foundation (Figure~\ref{fig:my_image_1}), supporting polygon review, area sorting, review-status tracking, COCO/VIA export, and overlay generation. Because reviewed cases, spatial labels, and disposition metadata are stored together, the same interface can support onboarding workflows in which trainees compare decisions against expert-reviewed reference cases, although that use is outside the current benchmark.

\begin{figure}[!t]
    \centering
    \includegraphics[width=0.95\columnwidth]{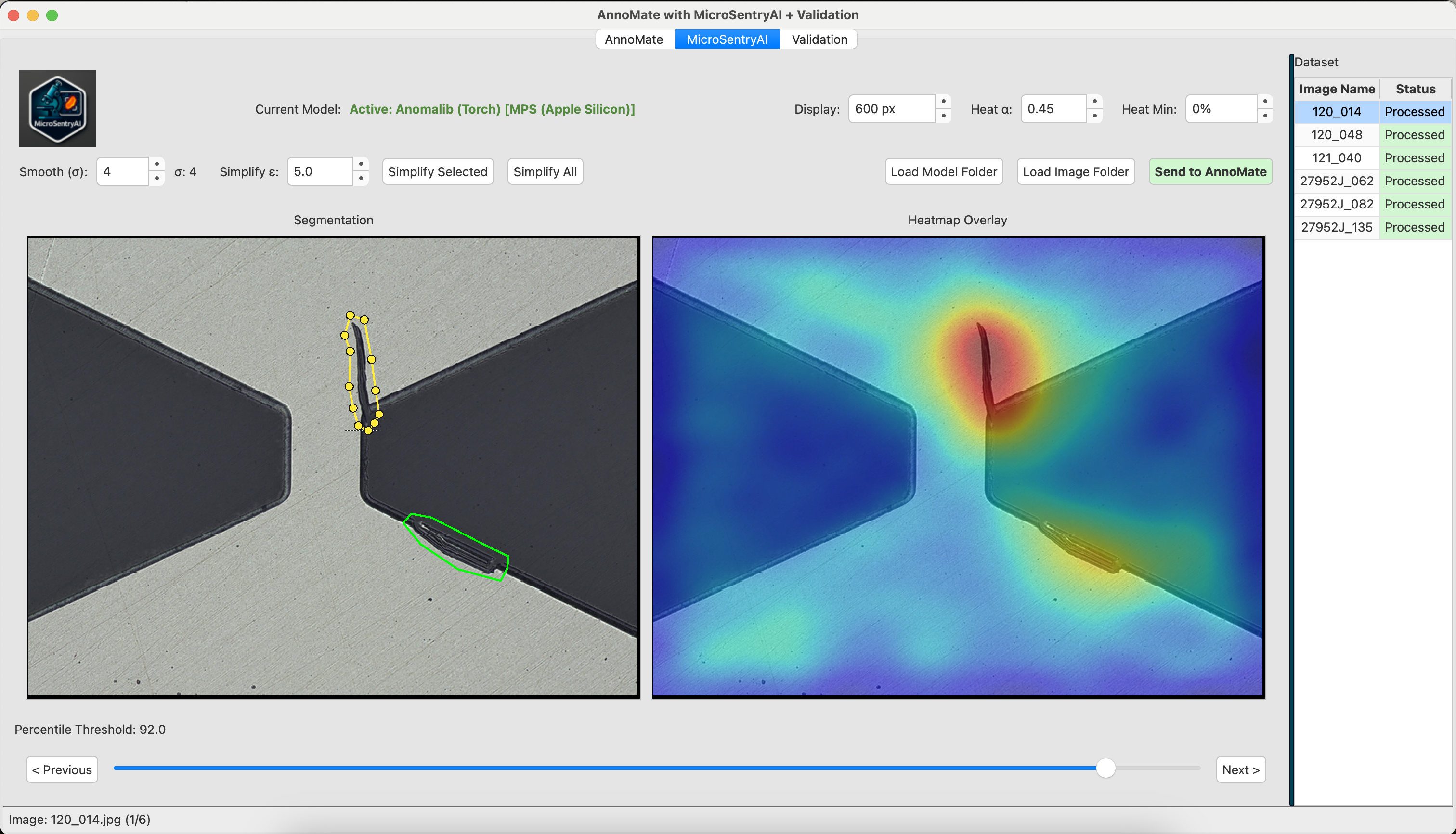}
    \caption{The \textit{MicroSentryAI} interface for anomaly inference and post-processing. A synchronized dual-view displays the predicted heatmap alongside editable polygon contours, enabling inspectors to localize candidate defects, refine or reject proposed regions, and confirm the overall image-level decision.}
    \label{fig:my_image_2}
    \vspace{-1mm}
\end{figure}

\textit{MicroSentryAI} adds inference and interactive post-processing (Figure~\ref{fig:my_image_2}). It loads trained weights, visualizes normalized anomaly maps, and exposes thresholding, smoothing, and contour controls. Heatmaps localize suspected defects, while SAM refines those regions into inspector-facing boundary proposals \cite{ravi2024sam2}. Inspectors can accept, reject, or adjust candidate segments and separately confirm the image-level decision before transferring accepted contours back to annotation. Here, SAM masks illustrate this refinement path rather than benchmark ground truth.

\subsection{Validation and Mask Evaluation}
\label{sec:evaluation}

\begin{figure}[t]
    \centering
    \includegraphics[width=\linewidth]{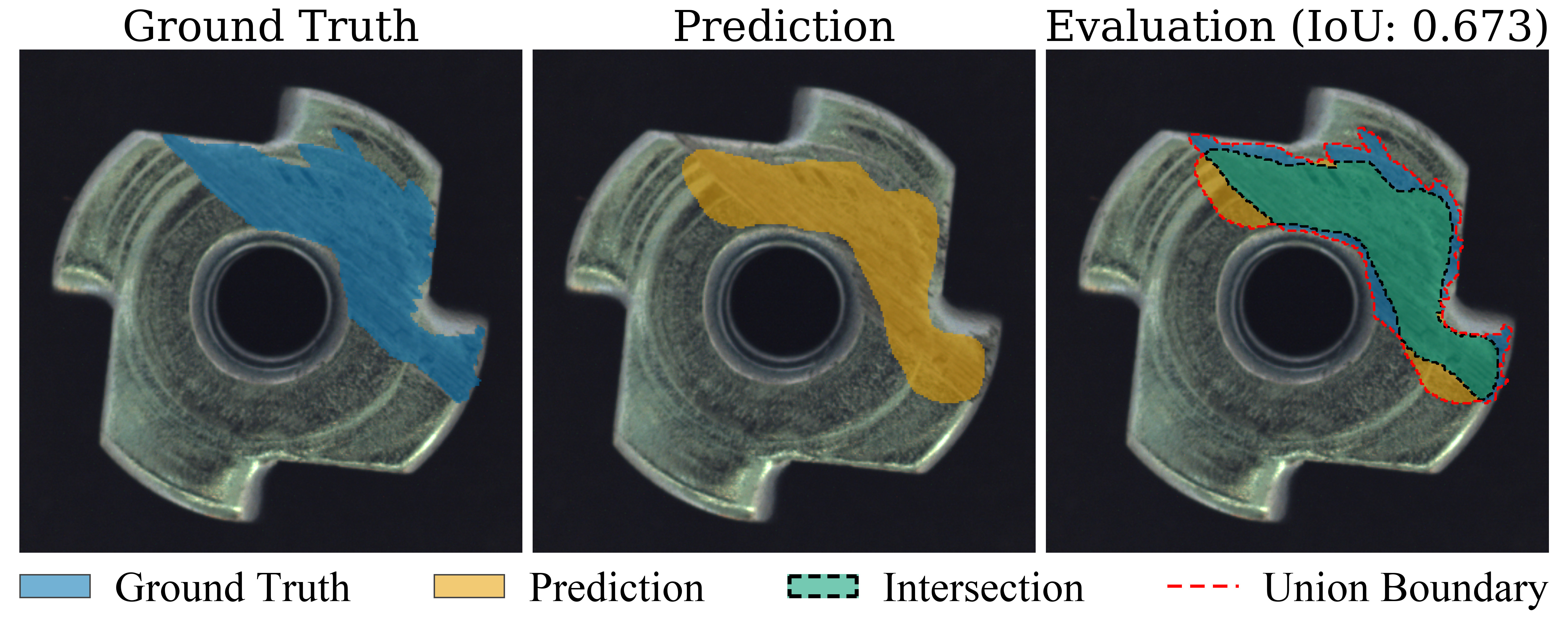}
    \caption{
        Mask-overlap visualization evaluating a PatchCore
        \cite{roth2022totalrecallindustrialanomaly}
        model's spatial accuracy (pixel AUROC: 0.989) on a metal nut from MVTec.
        The intersection of ground truth and prediction masks yields an IoU of 0.673.
    }
    \label{fig:evaluation_mask_analysis}
    \vspace{-1mm}
\end{figure}
\begin{figure}[t]
    \centering
    \includegraphics[width=\linewidth]{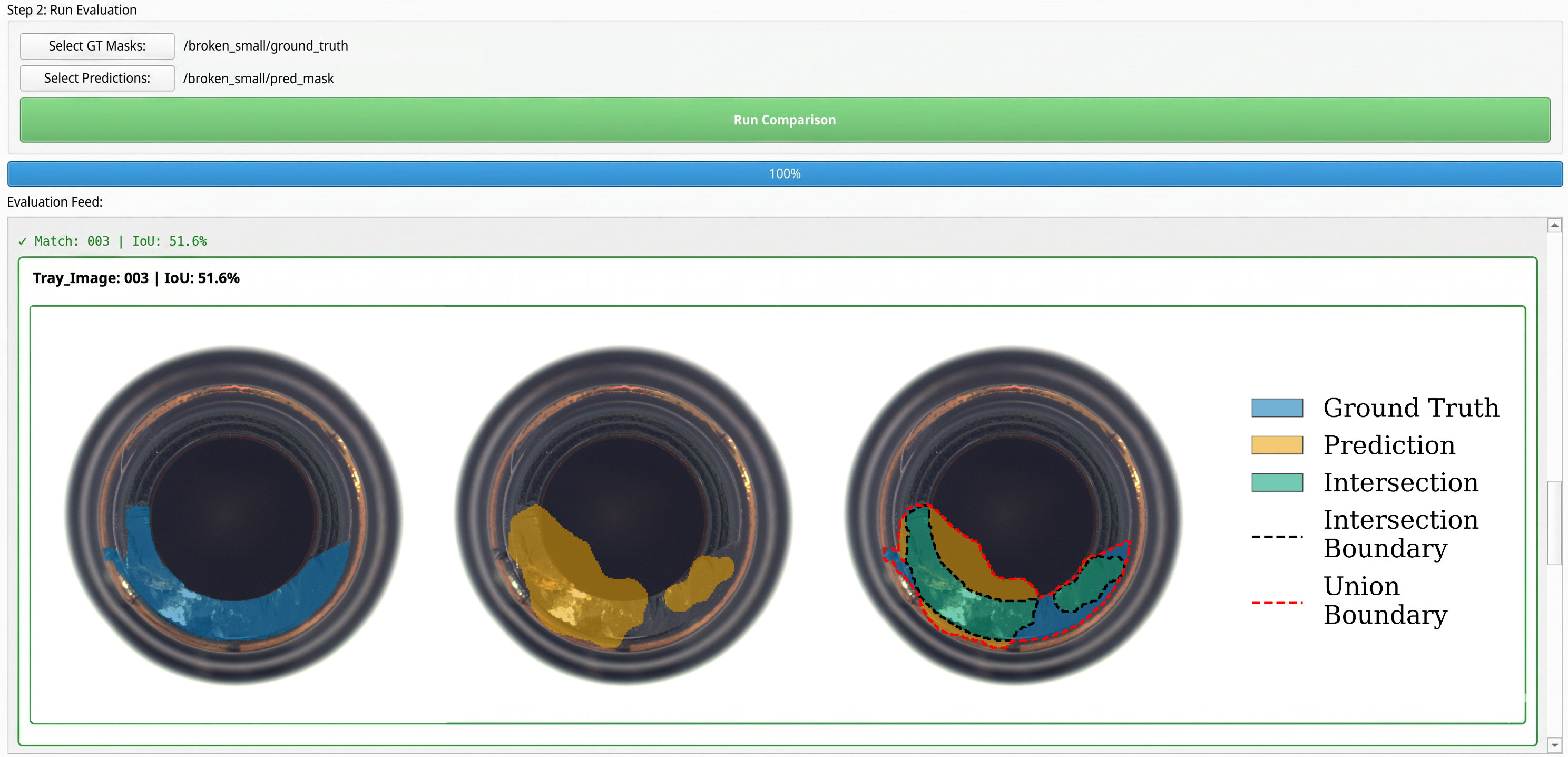}
    \caption{
        The validation interface for quantitative mask evaluation.
        It visually overlays ground-truth and predicted masks to calculate
        spatial accuracy metrics, such as Intersection over Union (IoU),
        directly within the workflow, in this case, for a bottle from
        MVTec.
    }
    \label{fig:my_image_3}
    \vspace{-1mm}
\end{figure}

The validation module closes the loop by evaluating masks inside the same workflow rather than as disconnected post-processing. Figure~\ref{fig:evaluation_mask_analysis} shows mask overlap, and Figure~\ref{fig:my_image_3} shows the evaluation interface using MVTec examples.\footnote{
    MVTec data~\cite{bergmann2021mvtec} is used here because ground-truth defect masks for the BowTie dataset were not yet available at the time of article submission.
}
The engine converts \textit{AnnoMate} annotations to binary masks and reports IoU, precision, recall, and centroid distance for model-vs.-ground-truth validation or user-vs.-model comparison. These overlays help inspectors and researchers connect numeric scores with defect morphology and annotation ambiguity.

Together, the three modules operationalize the paper's core argument: annotation, inference, and quantitative validation are most effective when treated as connected parts of the same inspection process rather than as isolated tasks.

\section{Conclusion}
\label{sec:conclusion}

This paper examined the gap between anomaly detection benchmark performance and
operational reliability in a real-world industrial setting. Evaluating 19
unsupervised anomaly detection models on the BowTie dataset, a reflective
metallic manufacturing dataset with profile-specific appearance variation and
morphologically diverse defects, we found that no single architecture emerged as
uniformly robust. Preprocessing choices had strongly architecture-dependent
effects, and a consensus-based data audit indicated that nominal-data quality
further interacts with model sensitivity. EfficientAD, Dinomaly, PatchCore,
Reverse Distillation, and CS-Flow were the strongest candidates in the reported
experiments.

Motivated by these findings, we presented an initially deployed unified human-in-the-loop framework:
AnnoMate for guided annotation and review, MicroSentryAI for AI-assisted defect
localization and segmentation review, and a Validation Engine for mask-based
comparison against expert references where ground truth is available. Together,
the modules replace a prior manual visual inspection and documentation workflow
with coupled curation, inference, and quantitative assessment. MicroSentryAI
supports heatmap-guided localization, SAM-refined candidates, inspector
acceptance or rejection, boundary adjustment, and image-level confirmation;
the Validation Engine reports IoU, precision, recall, and centroid distance.
The same environment also preserves localized decision rationale and can support
inspector consistency and onboarding through expert-reviewed cases, although that
use was not evaluated here.

Because the benchmark uses one real-world manufacturing dataset and lacked
pixel-level BowTie masks at submission, rankings should be read as a BowTie-specific
stress test rather than general capability claims. Future work will complete
BowTie masks, rerun spatial metrics, measure inspection-time, consistency, and
onboarding effects, and test additional profiles and defect modalities.

\bibliographystyle{ieeetr}
\bibliography{sections/references}

@misc{oceans11,
  author =        {{Los Alamos National Laboratory}},
  howpublished =  {\url{https://oceans11.lanl.gov}},
  note =          {Accessed: Apr. 13, 2026},
  title =         {{Oceans 11 Data Repository}},
  year =          {2026},
}

@misc{annomate-microsentryai-workflow,
  author =        {{Los Alamos National Laboratory}},
  howpublished =
  {\url{https://github.com/lanl/annomate-microsentryai-workflow}},
  note =          {GitHub repository, accessed April 2026},
  title =         {AnnoMate with MicroSentryAI: Human-in-the-loop
                   AIML-powered inspection tools},
  year =          {2026},
}

@article{SAHAR2023100803,
  author =        {Tayyaba Sahar and others},
  journal =       {{Results Eng.}
                       \href{https://doi.org/10.1016/j.rineng.2022.100803}{doi:
  10.1016/j.rineng.2022.100803}},
  title =         {Anomaly detection in laser powder bed fusion using
                   machine learning: A review},
  volume =        {17},
  year =          {2023},
  doi =           {10.1016/j.rineng.2022.100803},
}

@article{BENMOUSSAT201368,
  author =        {M.S. Benmoussat and M. Guillaume and Y. Caulier and
                   K. Spinnler},
  journal =       {{Infrared Phys. \& Tech.}
  \href{https://doi.org/10.1016/j.infrared.2013.07.007}{doi:
  10.1016/j.infrared.2013.07.007}},
  pages =         {68-80},
  title =         {Automatic metal parts inspection: Use of
                   thermographic images and anomaly detection
                   algorithms},
  volume =        {61},
  year =          {2013},
  doi =           {10.1016/j.infrared.2013.07.007},
}

@article{KAJI2022624,
  author =        {Farzaneh Kaji and others},
  journal =       {{J. Manuf. Process.}
                       \href{https://doi.org/10.1016/j.jmapro.2022.06.046}{doi:
  10.1016/j.jmapro.2022.06.046}},
  pages =         {624-637},
  title =         {A deep-learning-based in-situ surface anomaly
                   detection methodology for laser directed energy
                   deposition via powder feeding},
  volume =        {81},
  year =          {2022},
  doi =           {10.1016/j.jmapro.2022.06.046},
}

@inproceedings{dutta2019via,
  author =        {Abhishek Dutta and Andrew Zisserman},
  booktitle =     {{ACM Multimedia}
                   \href{https://doi.org/10.1145/3343031.3350535}{doi:
                   10.1145/3343031.3350535}},
  title =         {The VIA Annotation Software for Images, Audio and
                   Video},
  year =          {2019},
  doi =           {10.1145/3343031.3350535},
}

@article{russell2008labelme,
  author =        {Bryan C. Russell and Antonio Torralba and
                   Kevin P. Murphy and William T. Freeman},
  journal =       {{Int. J. Comput. Vis.}
                   \href{https://doi.org/10.1007/s11263-007-0090-8}{doi:
                   10.1007/s11263-007-0090-8}},
  number =        {1--3},
  pages =         {157--173},
  title =         {LabelMe: A Database and Web-Based Tool for Image
                   Annotation},
  volume =        {77},
  year =          {2008},
  doi =           {10.1007/s11263-007-0090-8},
}

@article{chandola2009anomaly,
  address =       {New York, NY, USA},
  author =        {Chandola, Varun and Banerjee, Arindam and
                   Kumar, Vipin},
  journal =       {ACM Comput. Surv.},
  month =         jul,
  note =          {\href{https://doi.org/10.1145/1541880.1541882}{doi:
                   10.1145/1541880.1541882}},
  number =        {3},
  publisher =     {Association for Computing Machinery},
  title =         {Anomaly detection: A survey},
  volume =        {41},
  year =          {2009},
  doi =           {10.1145/1541880.1541882},
  issn =          {0360-0300},
  url =           {https://doi.org/10.1145/1541880.1541882},
}

@article{pang2021anomaly,
  address =       {New York, NY, USA},
  author =        {Pang, Guansong and Shen, Chunhua and Cao, Longbing and
                   Hengel, Anton Van Den},
  journal =       {ACM Comput. Surv.},
  month =         mar,
  number =        {2},
  publisher =     {Association for Computing Machinery},
  title =         {{Deep Learning for Anomaly Detection: A Review}
                   \href{https://doi.org/10.1145/3439950}{doi:
                   10.1145/3439950}},
  year =          {2021},
  doi =           {10.1145/3439950},
  issn =          {0360-0300},
  url =           {https://doi.org/10.1145/3439950},
}

@inproceedings{akcay2022anomalib,
  author =        {Akcay, Samet and others},
  booktitle =     {ICIP},
  pages =         {1706-1710},
  title =         {{Anomalib: A Deep Learning Library for Anomaly
                   Detection}
  \href{https://doi.org/10.1109/ICIP46576.2022.9897283}{doi:
  10.1109/ICIP46576.2022.9897283}},
  year =          {2022},
  doi =           {10.1109/ICIP46576.2022.9897283},
}

@misc{zhang2023makesgooddataaugmentation,
  author =        {Lingrui Zhang and others},
  note =
  {\href{https://arxiv.org/abs/2304.03294}{arXiv:2304.03294}},
  title =         {What makes a good data augmentation for few-shot
                   unsupervised image anomaly detection?},
  year =          {2023},
  url =           {https://arxiv.org/abs/2304.03294},
}

@article{bergmann2021mvtec,
  author =        {Bergmann, Paul and Batzner, Kilian and
                   Fauser, Michael and Sattlegger, David and
                   Steger, Carsten},
  journal =       {Int. J. Comput. Vis.},
  note =              {\href{https://doi.org/10.1007/s11263-020-01400-4}{doi:
  10.1007/s11263-020-01400-4}},
  number =        {4},
  pages =         {1038--1059},
  title =         {The {MVTec} Anomaly Detection Dataset: {A}
                   Comprehensive Real-World Dataset for Unsupervised
                   Anomaly Detection},
  volume =        {129},
  year =          {2021},
}

@article{Liu_2024,
  author =        {Liu, Jiaqi and others},
  journal =       {Mach. Intell. Res.},
  note =          {\href{https://doi.org/10.1007/s11633-023-1459-z}{doi:
                   10.1007/s11633-023-1459-z}},
  number =        {1},
  pages =         {104-135},
  title =         {Deep Industrial Image Anomaly Detection: A Survey},
  volume =        {21},
  year =          {2024},
}

@inproceedings{xie2024imiadindustrialimageanomaly,
  author =        {Xie, Guoyang and others},
  booktitle =     {CVPR},
  note =
  {\href{https://arxiv.org/abs/2301.13359}{arXiv:2301.13359}},
  title =         {{IM-IAD}: Industrial Image Anomaly Detection
                   Benchmark in Manufacturing},
  year =          {2024},
  url =           {https://arxiv.org/abs/2301.13359},
}

@inproceedings{batzner2024efficientadaccuratevisualanomaly,
  author =        {Batzner, Kilian and Heckler, Lars and Konig, Rebecca},
  booktitle =     {WACV},
  note =              {\href{https://doi.org/10.1109/WACV57701.2024.00020}{doi:
  10.1109/WACV57701.2024.00020}},
  pages =         {127-137},
  title =         {{EfficientAD: Accurate Visual Anomaly Detection at
                   Millisecond-Level Latencies}},
  year =          {2024},
}

@inproceedings{deng2022anomalydetectionreversedistillation,
  author =        {Deng, Hanqiu and Li, Xingyu},
  booktitle =     {CVPR},
  note =              {\href{https://doi.org/10.1109/CVPR52688.2022.00951}{doi:
  10.1109/CVPR52688.2022.00951}},
  pages =         {9737--9746},
  title =         {Anomaly Detection via Reverse Distillation from
                   One-Class Embedding},
  year =          {2022},
}

@inproceedings{wang2021studentteacherfeaturepyramidmatching,
  author =        {Wang, Guodong and Han, Shumin and Ding, Errui and
                   Huang, Di},
  booktitle =     {BMVC},
  note =          {\href{https://doi.org/10.5244/C.35.349}{doi:
                   10.5244/C.35.349}},
  title =         {Student-Teacher Feature Pyramid Matching for Anomaly
                   Detection},
  year =          {2021},
  url =           {https://arxiv.org/abs/2103.04257},
}

@inproceedings{roth2022totalrecallindustrialanomaly,
  author =        {Roth, Karsten and Pemula, Latha and Zepeda, Joaquin and
                   Scholkopf, Bernhard and Brox, Thomas and
                   Gehler, Peter},
  booktitle =     {CVPR},
  note =              {\href{https://doi.org/10.1109/CVPR52688.2022.01392}{doi:
  10.1109/CVPR52688.2022.01392}},
  pages =         {14318-14328},
  title =         {{Towards Total Recall in Industrial Anomaly
                   Detection}},
  year =          {2022},
}

@inproceedings{defard2020padimpatchdistributionmodeling,
  author =        {Defard, Thomas and Setkov, Aleksandr and
                   Loesch, Angelique and Audigier, Romaric},
  booktitle =     {ICPR Workshops},
  note =              {\href{https://doi.org/10.1007/978-3-030-68799-1_35}{doi:
  10.1007/978-3-030-68799-1\_35}},
  pages =         {475-489},
  title =         {{PaDiM}: A Patch Distribution Modeling Framework for
                   Anomaly Detection and Localization},
  year =          {2021},
}

@article{lee2022cfacoupledhyperspherebasedfeatureadaptation,
  author =        {Lee, Sungwook and Lee, Seunghyun and
                   Song, Byung Cheol},
  journal =       {IEEE Access},
  note =              {\href{https://doi.org/10.1109/ACCESS.2022.3193699}{doi:
  10.1109/ACCESS.2022.3193699}},
  pages =         {78446-78454},
  title =         {{CFA}: Coupled-hypersphere-based Feature Adaptation
                   for Target-Oriented Anomaly Localization},
  volume =        {10},
  year =          {2022},
  url =           {https://ieeexplore.ieee.org/stamp/stamp.jsp?
                  arnumber=9839549},
}

@inproceedings{damm2025anomalydinoboostingpatchbasedfewshot,
  author =        {Damm, Simon and Laszkiewicz, Mike and
                   Lederer, Johannes and Fischer, Asja},
  booktitle =     {WACV},
  note =              {\href{https://doi.org/10.1109/WACV61041.2025.00136}{doi:
  10.1109/WACV61041.2025.00136}},
  pages =         {1319-1329},
  title =         {{AnomalyDINO: Boosting Patch-based Few-Shot Anomaly
                   Detection with DINOv2}},
  year =          {2025},
}

@inproceedings{rolih2024supersimplenetunifyingunsupervisedsupervised,
  author =        {Rolih, Bla{\v{z}} and Fu{\v{c}}ka, Matic and
                   Sko{\v{c}}aj, Danijel},
  booktitle =     {Pattern Recognit.},
  note =
  {\href{https://arxiv.org/abs/2408.03143}{arXiv:2408.03143}},
  pages =         {47-65},
  title =         {{SuperSimpleNet: Unifying Unsupervised and Supervised
                   Learning for Fast and Reliable Surface Defect
                   Detection}},
  year =          {2025},
}

@misc{yu2021fastflowunsupervisedanomalydetection,
  author =        {Yu, Jiawei and Zheng, Ye and Wang, Xiang and Li, Wei and
                   Wu, Yushuang and Zhao, Rui and Wu, Liwei},
  note =
  {\href{https://arxiv.org/abs/2111.07677}{arXiv:2111.07677}},
  title =         {{FastFlow: Unsupervised Anomaly Detection and
                   Localization via 2D Normalizing Flows}},
  year =          {2021},
}

@inproceedings{gudovskiy2021cflowadrealtimeunsupervisedanomaly,
  author =        {Gudovskiy, Denis and Ishizaka, Shun and
                   Kozuka, Kazuki},
  booktitle =     {WACV},
  note =              {\href{https://doi.org/10.1109/WACV51458.2022.00188}{doi:
  10.1109/WACV51458.2022.00188}},
  pages =         {1819-1828},
  title =         {{CFLOW-AD: Real-Time Unsupervised Anomaly Detection
                   with Localization via Conditional Normalizing Flows}},
  year =          {2022},
}

@inproceedings{rudolph2021fullyconvolutionalcrossscaleflowsimagebased,
  author =        {Rudolph, Marco and Wehrbein, Tom and Rosenhahn, Bodo and
                   Wandt, Bastian},
  booktitle =     {WACV},
  note =              {\href{https://doi.org/10.1109/WACV51458.2022.00189}{doi:
  10.1109/WACV51458.2022.00189}},
  pages =         {1829-1838},
  title =         {{Fully Convolutional Cross-Scale-Flows for
                   Image-based Defect Detection}},
  year =          {2022},
}

@article{tailanian2024uflowushapednormalizingflow,
  author =        {Tailanian, Mat{\'i}as and Pardo, {\'A}lvaro and
                   Mus{\'e}, Pablo},
  journal =       {J. Math. Imaging Vis.},
  note =              {\href{https://doi.org/10.1007/s10851-024-01193-y}{doi:
  10.1007/s10851-024-01193-y}},
  number =        {4},
  pages =         {678-696},
  title =         {{U-Flow}: A U-Shaped Normalizing Flow for Anomaly
                   Detection with Unsupervised Threshold},
  volume =        {66},
  year =          {2024},
}

@inproceedings{guo2025dinomalyphilosophymulticlassunsupervised,
  author =        {Guo, Jia and others},
  booktitle =     {CVPR},
  note =
  {\href{https://openaccess.thecvf.com/content/CVPR2025/html/Guo_Dinomaly_The_Less_Is_More_Philosophy_in_Multi-Class_Unsupervised_Anomaly_CVPR_2025_paper.html}{paper:
  CVPR OA}},
  pages =         {20405-20415},
  title =         {{Dinomaly: The Less Is More Philosophy in Multi-Class
                   Unsupervised Anomaly Detection}},
  year =          {2025},
}

@inproceedings{zavrtanik2021draemdiscriminativelytrained,
  author =        {Zavrtanik, Vitjan and Kristan, Matej and
                   Skocaj, Danijel},
  booktitle =     {ICCV},
  note =              {\href{https://doi.org/10.1109/ICCV48922.2021.00822}{doi:
  10.1109/ICCV48922.2021.00822}},
  pages =         {8310-8319},
  title =         {{DRÆM – A discriminatively trained reconstruction
                   embedding for surface anomaly detection}},
  year =          {2021},
}

@inproceedings{zavrtanik2022dsrdualsubspace,
  author =        {Zavrtanik, Vitjan and Kristan, Matej and
                   Sko{\v{c}}aj, Danijel},
  booktitle =     {ECCV},
  note =
  {\href{https://doi.org/10.1007/978-3-031-19821-2%5F31}{doi:
  10.1007/978-3-031-19821-2\_31}},
  pages =         {539-554},
  title =         {{DSR} -- A Dual Subspace Re-Projection Network for
                   Surface Anomaly Detection},
  year =          {2022},
}

@inproceedings{akcay2018ganomalysemisupervisedanomalydetection,
  author =        {Akcay, Samet and Atapour-Abarghouei, Amir and
                   Breckon, Toby P.},
  booktitle =     {ACCV},
  title =         {{GANomaly: Semi-Supervised Anomaly Detection via
                   Adversarial Training}
                   \href{https://arxiv.org/abs/1805.06725}{[arXiv]}},
  year =          {2018},
  url =           {https://arxiv.org/abs/1805.06725},
}

@misc{ndiour2022frefastmethodanomaly,
  author =        {Ndiour, Ibrahima and Ahuja, Nilesh and Genc, Utku and
                   Tickoo, Omesh},
  title =         {{FRE: A Fast Method For Anomaly Detection And
                   Segmentation}
                   \href{https://arxiv.org/abs/2211.12650}{[arXiv]}},
  year =          {2022},
  url =           {https://arxiv.org/abs/2211.12650},
}

@misc{ahuja2019probabilisticmodelingdeepfeatures,
  author =        {Ahuja, Nilesh A. and Ndiour, Ibrahima and
                   Kalyanpur, Trushant and Tickoo, Omesh},
  note =
  {\href{https://arxiv.org/abs/1909.11786}{arXiv:1909.11786}},
  title =         {{Probabilistic Modeling of Deep Features for
                   Out-of-Distribution and Adversarial Detection}},
  year =          {2019},
}

@misc{experiments-reproducibility,
  author =        {{Los Alamos National Laboratory}},
  howpublished =  {\url{https://tinyurl.com/placeholder}},
  note =          {Accessed: 2026},
  title =         {Repository for Reproducibility for {ICMLA} 2026},
  year =          {2026},
}

@inproceedings{ravi2024sam2,
  author =        {Ravi, Nikhila and others},
  booktitle =     {ICLR},
  note =              {\href{https://openreview.net/forum?id=Ha6RTeWMd0}{paper:
  OpenReview}},
  title =         {{SAM 2: Segment Anything in Images and Videos}},
  year =          {2025},
}

\end{document}